\documentclass[referee,sn-apa]{sn-jnl}

\usepackage{graphicx}%
\usepackage{multirow}%
\usepackage{amsmath,amssymb,amsfonts}%
\usepackage{amsthm}%
\usepackage{mathrsfs}%
\usepackage[title]{appendix}%
\usepackage{xcolor}%
\usepackage{textcomp}%
\usepackage{manyfoot}%
\usepackage{booktabs}%
\usepackage{algorithm}%
\usepackage{algorithmicx}%
\usepackage{algpseudocode}%
\usepackage{listings}%
\usepackage{natbib}
\usepackage{subcaption}

\begin{document}

\title[Article Title]{Visual-inertial state estimation based on Chebyshev polynomial optimization
}


\author[1]{\fnm{Hongyu} \sur{Zhang}}\email{hongyuzhang@sjtu.edu.cn}

\author[1]{\fnm{Maoran} \sur{Zhu}}\email{zhumaoran@sjtu.edu.cn}

\author[1]{\fnm{Qi} \sur{Cai}}\email{qicaiCN@gmail.com}

\author*[1]{\fnm{Yuanxin} \sur{Wu}}\email{yuanx\_wu@hotmail.com}

\affil*[1]{\orgdiv{Shanghai Key Laboratory of Navigation and Location Based Services}, \orgname{Shanghai Jiao Tong University}, \orgaddress{
\city{Shanghai}, \postcode{200240}, 
\country{China}}}




\abstract{ This paper proposes an innovative state estimation method for visual-inertial fusion based on Chebyshev polynomial optimization. Specifically, the pose is modeled as a Chebyshev polynomial of a certain order, and its time derivatives are used to calculate linear acceleration and angular velocity, which, along with inertial measurements, constitute dynamic constraints. This is coupled with a visual measurement model to construct a visual-inertial bundle adjustment formulation. Simulation and public dataset experiments show that the proposed method has better accuracy than the discrete-form preintegration method.}




\maketitle

\section{Introduction}\label{sec1}

Visual-inertial Navigation System (VINS) \citep{bib1} has a wide range of application scenarios, including but not limited to autonomous robot navigation, autonomous driving, and virtual reality. VINS employs sequences of images and outputs from the Inertial Measurement Unit (IMU)—such as angular velocities and specific forces—to estimate the attitude, velocity, and position of the carrier. Owing to the complementary nature of IMU and camera, visual-inertial fusion results in more stable pose estimation compared to the use of a single sensor \citep{bib2}. However, VINS is confronted with the challenge of cumulative motion drift over time due to the absence of global information. The core of VINS is the visual-inertial fusion state estimation algorithm. Early works \citep{bib2,bib3,bib4} primarily employed filtering methods, among which the most common is based on Extended Kalman Filter (EKF) \citep{bib5}. In this approach, IMU measurements are used for state propagation, while images are used for state updating. A representative algorithm is the Multi-State Constraint Kalman Filter (MSCKF) \citep{bib2}, which utilizes inertial dynamics equations for state propagation and projects the visual measurement residuals onto the null space of the feature Jacobian matrix for state updating. This avoids the estimation of a large number of feature points and thus reduces computational costs. Additionally, other filtering algorithms such as Invariant Extended Kalman Filter (IEKF) \citep{bib6} and Unscented Kalman Filter (UKF) \citep{bib7} have also been applied to address the challenges in visual-inertial fusion.

Filter-based VINS exhibits high computational efficiency, yet during state updates, camera measurements must undergo linearization, consequently diminishing the precision of state estimation\citep{bib1}. Over the past decade, optimization-based algorithms\citep{bib8,bib9,bib10,bib11} have sought to mitigate the errors induced by linearization. Typically, these algorithms integrate IMU measurements between two image frames to establish relative constraints. Standard IMU integration relies upon the poses to be estimated, which are subject to alteration with each iteration during the optimization process, necessitating recalculation of the IMU integration, and thus incurring substantial computational costs. Recent work(e.g., \cite{bib9,bib10,bib11}) usually employs the method of IMU preintegration \citep{bib12}, which integrates IMU measurements in the local frame to generate inter-frame relative pose constraints, and uses linearized bias to update these constraints in order to avoid re-integration. However, IMU preintegration, as a relative constraint, loses the quasi-Gaussian nature of the original measurements, and the linearization of the bias also compromises the precision of the functional model to some extent.

The algorithms discussed above are all based on discrete time, directly estimating the poses of each camera keyframe. However, discrete-time methods require specialized algorithms to handle the asynchronous measurements from multiple sensors \citep{bib13}, which is not conducive to system scalability. In recent years, several continuous-time visual-inertial fusion algorithms have been proposed \citep{bib14,bib15,bib16,bib17,bib18}. First, continuous-time poses can be sampled at any moment, facilitating the fusion of asynchronous measurements from multiple sensors, and are also advantageous for state estimation with sensors that output a continuous data stream, such as rolling shutter cameras. Second, continuous-time poses do not require the estimation of poses at each sensor measurement point, and the state dimension depends on the polynomial order of the pose representation, facilitating the fusion of sensors with different sampling rates. Considering local support and analytical derivatives and integrals, they usually employed (cumulative) B-splines \citep{bib19} to represent the continuous poses, but B-splines are not computation-friendly for real-time calculations \citep{bib16}.

Inspired by the collocation method, we quite recently come up with a continuous-time state estimation method by the Chebyshev polynomial optimization for general state estimation problems \citep{bib20}, and applies it to inertial-magnetic attitude estimation \citep{bib33}. This paper will further leverage the concept of collocation to construct a visual-inertial fusion state estimation algorithm based on polynomial optimization, transforming pose estimation into a Chebyshev coefficient optimization problem. Specifically, the attitude and velocity are represented by Chebyshev polynomials, with the position obtained through the analytical integration of velocity. The unknown coefficients are then determined by minimizing the weighted residuals of initial conditions, dynamics and measurements, thereby obtaining continuous poses. This method directly utilizes the original IMU measurements to construct the objective function, avoiding the linearization issues prevalent in filtering methods, such as those described by  \cite{bib2} and \cite{bib3}. Compared to discrete-time methods employing IMU preintegration, like those presented by \cite{bib9} and \cite{bib11}, it preserves the quasi-Gaussian nature of the original measurements. Additionally, the continuous-time format, similar to other continuous-time methods (e.g., \cite{bib15,bib17}), facilitates the fusion of multi-frequency asynchronous sensors. Moreover, the employment of Chebyshev polynomials ensures high accuracy and efficiency. Simulation and experimental results demonstrate that the method proposed in this paper exhibits better accuracy compared to the classic preintegration methods.

The main contribution of this work rests on a novel framework for continuous-time visual-inertial state estimation based on Chebyshev polynomial optimization. Compared with the well-known IMU preintegration, the proposed estimation method does not introduce any approximation of the nonlinear dynamics and measurements, thus achieving remarkably improved accuracy. The remainder of this paper is structured as follows: Section \ref{sec2} provides an overview of the visual-inertial measurement models and introduces the state estimation framework based on Chebyshev polynomial optimization. Section \ref{sec3} focuses on evaluating the proposed algorithm, showcasing its improved accuracy compared to traditional IMU preintegration. Finally, Section \ref{sec4} presents discussions and draws conclusions.

\begin{table}[ht]
  \centering
  \caption{Glossary of notations}
  \begin{tabular}{ll}
    \toprule
    \textbf{Symbol} & \textbf{Meaning} \\
    \midrule
    w & World frame \\
    n & North-Up-East(NUE) frame \\
    b & Body(IMU) frame \\
    c & Camera frame \\
    $q_B^{A}(C_B^{A})$ & Rotation of frame A relative to frame B \\
    $v_{B A}^{C}$ & Frame A's velocity relative to frame B, expressed in frame C \\
    $p_{B A}^{C}$ & Frame A's position relative frame B, expressed in frame C \\
    $\hat{x}$  & Estimated value of x \\
    $\tilde{x}$   & error-contaminated value of x \\
    $e_i$ & The $i$th column of $I_3$ \\
    \bottomrule
  \end{tabular}
  \label{tab-notations}
\end{table}

\section{Methodology}\label{sec2}
Before describing the overall algorithm, The coordinate frames and some notations involved are clarified in Table \ref{tab-notations}. The IMU frame and body frame are equivalent by default in the following discussion. 
\subsection{Visual-inertial measurement model}\label{sec2.1}
\subsubsection{IMU measurement model}\label{sec2.1.1}
Define the IMU state:
\begin{equation}
x_I=\begin{bmatrix} {q_w^{b}}^T &  {v_{w b}^{w}}^T &  {p_{w b}^{w}}^T &  b_a^T &  b_g^T\end{bmatrix}^T
\label{eq1}
\end{equation}
where $I$ represents the IMU, $w$ represents the world frame, $b$ represents the body frame, $q_w^{b}$, $v_{w b}^{w}$ and $p_{w b}^{w}$ respectively represent the attitude quaternion, the velocity, and the position of the body frame relative to the world frame, and $b_a$ and $b_g$ are the accelerometer bias and gyroscope bias, respectively.

In consumer-grade inertial navigation applications, the influence of Earth's rotation is typically not considered. The continuous-time state update equation for the IMU is as follows \citep{bib21}
\begin{equation}
\dot{q}_w^b=\frac{1}{2} q_w^b\circ
\begin{bmatrix}
0 \\ \omega_{w b}^b
\end{bmatrix}
\label{eq2}
\end{equation}
\begin{equation}
\dot{v}_{w b}^w=f_{w b}^w+g^w={C_w^{b}}^T f_{w b}^b+g^w \label{eq3} 
\end{equation}
\begin{equation}
\dot{p}_{w b}^w=v_{w b}^w \label{eq4} 
\end{equation}
where $C_w^b$ denotes the attitude matrix or rotation matrix of the body frame relative to the world frame (which can be derived from $q_w^b$ ), $g^w$ is the gravity in the world frame, and $\omega_{wb}^b$ and $f_{wb}^b$ are the true values of the angular velocity and specific force, respectively.
The gyroscope angular velocity measurement $\tilde{\omega}_{w b}^b$ and accelerometer specific force measurement $\tilde{f}_{w b}^b$ are given by 
\begin{equation}
    \tilde{\omega}_{w b}^b=\omega_{w b}^b+b_g+n_g \label{eq5}
\end{equation}
\begin{equation}
    \tilde{f}_{w b}^b=f_{w b}^b+b_a+n_a
    \label{eq6}
\end{equation}
where it is assumed that the noise of the gyroscope and accelerometer follows a zero-mean Gaussian distribution, namely $n_g \sim N\left(0, R_g\right)$, $n_a \sim N\left(0, R_a\right)$. The biases follow a random walk model 
\begin{align}
    \dot{b}_g=n_{b_g} \nonumber \\
    \dot{b}_a=n_{b_a} \label{eq7}
\end{align}
where $n_{b_g} \sim N\left(0, R_{b_g}\right)$ and $n_{b_a} \sim N\left(0, R_{b_a}\right)$ are the random walk noises for the gyroscope bias and accelerometer bias, respectively.

By substituting (\ref{eq5}) into (\ref{eq2}) and multiplying both sides by the conjugate of $q_w^b$, the angular velocity measurement can be expressed as a function of the attitude quaternion and gyroscope bias 
\begin{equation}
    \tilde{\omega}_{w b}^b=2\left[{q_w^{b}}^* \circ \dot{q}_w^b\right]_{2: 4}+b_g+n_g \label{eq8}
\end{equation}
where the operator $[\cdot ]_{2: 4}$ extracts the second to the fourth rows of a matrix. By substituting (\ref{eq6}) into (\ref{eq3}) and multiplying both sides by $C_w^b$, the specific force measurement can be expressed as a function of the attitude, velocity, and accelerometer bias
\begin{equation}
    \tilde{f}_{w b}^b=C_w^b\left(\dot{v}_{w b}^w-g^w\right)+b_a+n_a \label{eq9}
\end{equation}

\subsubsection{Visual measurement model} \label{sec2.1.2}
For the visual measurement model, this paper employs the classical bundle adjustment (BA) reprojection error \citep{bib22,bib23}. Let the 3D feature point in the world frame be $X_k^w$, and the noise-affected normalized projection coordinates in the $i$-th image be $\tilde{X}_k^{c_i}$. Then, the BA reprojection error for the feature point $X_k^w$ in the $i$-th image is typically defined as 
\begin{align}
& r_{B A}^{c_i}(k)=X_k^{c_i}-\tilde{X}_k^{c_i}=\frac{Y_{B A}^{c_i}(k)}{e_3^T Y_{B A}^{c_i}(k)}-\tilde{X}_k^{c_i} \nonumber \\
& Y_{B A}^{c_i}(k)=C_b^c\left(C_w^{b_i}\left(X_k^w-p_{w b_i}^w\right)\right)+p_{c b}^c \label{eq10}
\end{align}
where  $C_b^c$ and $p_{cb}^c$ are the extrinsic parameters between the camera and the IMU (which are known constants after calibration).

\subsection{Visual-inertial state estimation based on Chebyshev polynomial optimization}\label{sec2.2}
\subsubsection{Formulation of state estimation} \label{sec2.2.1}
Without the loss of the generality, consider the state estimation on the time interval $[t_0 \ t_M]$. Because the gyroscope and accelerometer biases are slowly changing, it is reasonable to treat them as constants over a short period. Therefore, the states to be estimated in the time interval include attitude, velocity, position, constant biases, and 3D feature points, denoted as $x(t) \triangleq[q_w^b(t)^T \  v_{w b}^w(t)^T \  p_{w b}^w(t)^T \  b_a^T \  b_g^T \  {X_1^w}^T \cdots {X_S^w}^T]^T$. Assume the initial IMU state is given as $\hat{x}_{I,0}=[\hat{q}_w^{b_0 T} \  \hat{v}_{w b_0}^{w T} \  \hat{p}_{w b_0}^{w T} \  \hat{b}_{a_0}^T \  \hat{b}_{g_0}^T]^T$, 
where $\hat{q}_w^{b_0}$, $\hat{v}_{w b_0}^w$, $\hat{p}_{w b_0}^{w}$, $\hat{b}_{a_0}$and $\hat{b}_{g_0}$ denote the attitude, velocity, position, accelerometer bias and gyroscope bias at the initial time $t_0$, respectively.

In the least squares sense\citep{bib24}, an optimal continuous-discrete state estimation is to minimize the measurement
residuals with respect to $x(t)$
\begin{equation}
    \min _{x(t)}\left(J_{x_{I, 0}}+J_a+J_g+J_c\right) \quad \text { s.t. }\left\|q_w^b(t)\right\|=1 \label{eq11}
\end{equation}
where the objective functions $J_{x_{I, 0}}$, $J_a$, $J_g$ and $J_c$, respectively denote the prior, specific force dynamics, angular velocity dynamics and visual measurement terms. They are explicitly given as follows
\begin{align}
& J_{x_{I, 0}}=e_{x_{I, 0}}^T e_{x_{I, 0}} \nonumber \\
& J_a=\int_{t_0}^{t_M} e_a^T(t) e_a(t) \mathrm{dt} \nonumber\\
& J_g=\int_{t_0}^{t_M} e_g^T(t) e_g(t) \mathrm{dt} \nonumber\\
& J_c=\sum_{k=1}^S \sum_{i \in S_k}\left(e_k^i\right)^T e_k^i  \label{eq12}
\end{align}
where $S$ is the number of 3D feature points within the interval, and $S_k$ denotes the set of images in which the corresponding feature points are observed. With the measurement models (\ref{eq8}), (\ref{eq9}) and (\ref{eq10}), the weighted residuals $e_{x_{I, 0}}$, $e_a(t)$, $e_g(t)$ and $e_k^i$ are given as 
\begin{align}
& e_{x_{I, 0}}=W_{x_{I, 0}}^T\left[\begin{array}{ccccc}2\left[q_w^{b_0^*} \circ \hat{q}_w^{b_0}\right]_{2: 4}^T & v_{w b_0}^{w T}-\hat{v}_{w b_0}^{w T} & p_{w b_0}^{w T}-\hat{p}_{w b_0}^{w T} & b_a^T-\hat{b}_{a_0}^T & b_g^T-\hat{b}_{g_0}^T\end{array}\right]^T \nonumber \\
& e_g=W_g^T\left(\tilde{\omega}_{w b}^b-2\left[q_w^{b^*} \circ \dot{q}_w^b\right]_{2: 4}-b_g\right) \nonumber \\
& e_a=W_a^T\left(\tilde{f}_{w b}^b-C_w^b \dot{v}_{w b}^w+C_w^b g^w-b_a\right)\nonumber \\
& e_k^i=W_c^T r_{B A}^{c_i}(k)  \label{eq13}
\end{align}
 The weight matrices are obtained from the Cholesky factorization of the inverse covariance matrix, namely $P_{x_{I, 0}}^{-1}=W_{x_{I, 0}} W_{x_{I, 0}}^T$, $R_a^{-1}=W_a W_a^T$, $R_g^{-1}=W_g W_g^T$, $R_c^{-1}=W_c W_c^T$. It should be noted that the
covariance matrix of the initial IMU state is defined as $P_{x_{I, 0}} \triangleq \operatorname{diag}[
P_{q_0} \  P_{v_0} \  P_{p_0} \  P_{b_a, 0} \  P_{b_g, 0}
]$, where $P_{q_0}$, $P_{v_0}$, $P_{p_0}$, $P_{b_a, 0}$ and $P_{b_g, 0}$ are the initial covariances of the attitude, velocity, position, accelerometer and gyroscope biases, respectively.

\subsubsection{State estimation based on Chebyshev polynomial optimization} \label{sec2.2.2}
To address the problem of continuous-time state estimation in (\ref{eq11}), we introduce the Chebyshev collocation method, representing attitude and velocity as finite-order Chebyshev polynomials, thereby converting the continuous-time state estimation problem into a constant parameter optimization problem.

The Chebyshev polynomial of the first kind is defined over the internal $[-1 \  1]$ by the recurrence relation as \citep{bib25}
\begin{align}
& F_0(\tau)=1, F_1(\tau)=\tau \nonumber \\
& F_{i+1}(\tau)=2 \tau F_i(\tau)-F_{i-1}(\tau) \text { for } i \geq 1  \label{eq14}
\end{align}
where $F_i(\tau)$ is the $i^{th}$-degree Chebyshev polynomial.The derivatives of the Chebyshev polynomials can be directly obtained by differentiating (\ref{eq14})
\begin{align}
& \dot{F}_0(\tau)=0, \dot{F}_1(\tau)=1 \nonumber \\
& \dot{F}_{i+1}(\tau)=2 F_i(\tau)+2 \tau \dot{F}_i(\tau)-\dot{F}_{i-1}(\tau) \text { for } i \geq 1 \label{15}
\end{align}
where $\dot{F}_i(\tau)$ is the derivative of the $i^{th}$-degree Chebyshev polynomial. And the integrated $i^{th}$-degree Chebyshev polynomial can be expressed as a linear combination of $(i+1)^{th}$-degree Chebyshev polynomial, given by \citep{bib26}
\begin{equation}
    G_{i,[-1, \tau]}=\int_{-1}^\tau F_i(\tau) d \tau=\left\{\begin{array}{l}
\left(\frac{F_{i+1}(\tau)}{2(i+1)}-\frac{F_{|i-1|}(\tau)}{2(i-1)}\right)-\frac{(-1)^i}{i^2-1} F_0(\tau), i \neq 1 \\
\frac{F_{i+1}(\tau)}{4}-\frac{F_0(\tau)}{4}, i=1
\end{array}\right. \label{eq16}
\end{equation}
In order to apply the Chebyshev polynomial, the optimization in (\ref{eq11}) on $t\in[t_0 \ t_M]$ needs to be mapped into $\tau \in[-1 \  1 ]$ by the affine transformation as follows
\begin{equation}
    \tau=\frac{2}{t_M-t_0} t-\frac{t_M+t_0}{t_M-t_0}
    \label{eq17}
\end{equation}
The attitude quaternion in the time interval is then approximated by a Chebyshev polynomial up to order $N_q$ as
\begin{equation}
    q_w^b(\tau) \approx \sum_{i=0}^{N_q} d_i F_i(\tau) \triangleq \mathbf{D F}(\tau) \label{eq18}
\end{equation}
where $d_i$ denotes the $i^{th}$-degree Chebyshev coefficient. The matrices $\mathbf{F}(\tau) \triangleq[F_0(\tau) \  F_1(\tau) \cdots F_{N_q}(\tau) ]^T$ and $\mathbf{D} \triangleq[ d_0 \  d_1 \cdots d_{N_q} ]$ are defined for compact denotation. The velocity in the time interval is approximated by a Chebyshev polynomial up to order $N_v$ as
\begin{equation}
    v_{w b}^w(\tau) \approx \sum_{i=0}^{N_v} k_i F_i(\tau) \triangleq \mathbf{K F}(\tau) \label{eq19}
\end{equation}
where $k_i$ denotes the $i^{th}$-degree Chebyshev coefficient and the matrix $\mathbf{K} \triangleq[ k_0 \  k_1 \cdots k_{N_v} ]$ is defined for compact denotation. The position in the time interval can be obtained by integrating the velocity over time
\begin{equation}
    p_{w b}^w(\tau)=p_{w b_0}^w+\frac{t_M-t_0}{2} \int_{-1}^\tau \sum_{i=0}^{N_v} k_i F_i(\tau) \mathrm{d} \tau=p_{w b_0}^w+\frac{t_M-t_0}{2} \sum_{i=0}^{N_v} k_i G_i(\tau) \label{eq20}
\end{equation}
The reason that the Chebyshev polynomial is selected as the basis function is for its high accuracy and efficacy in functional approximation. As a matter of fact, the Chebyshev polynomial is very close to the best polynomial approximation in the $\infty$-norm \citep{bib25}. The integral term $J_a$ and $J_g$ in (\ref{eq12}) can be numerically
solved by the Clenshaw-Curtis quadrature formula as \citep{bib25}
\begin{align}
& J_a \approx \sum_{i=0}^N w_i e_a^T\left(\tau_i\right) e_a\left(\tau_i\right) \nonumber\\
& J_g \approx \sum_{i=0}^N w_i e_g^T\left(\tau_i\right) e_g\left(\tau_i\right) \label{eq21}
\end{align}
where $\tau_i$ denotes the Chebyshev points
\begin{equation}
    \tau_i=-\cos (i \pi / N), i=0,1, \cdots, N \label{eq22}
\end{equation}
and $w_i$ denotes the weight that is determined by the integrals of the Lagrange polynomials \citep{bib27} and $N+1$ is the number of Chebyshev points. We see that (\ref{eq21}) requires the angular velocity and specific force at the Chebyshev points. In this paper, they are achieved by the way of the extended Floater and Hormann (EFH) interpolation method \citep{bib28}, which reconstructs the angular velocity and specific force from the equally-spaced time sampled angular velocity and specific force measurements, respectively. 

Following the commonly used strategy to handle the constraints in the collocation-based optimal control \citep{bib29}, the continuous unit quaternion constraint in (\ref{eq11}) is discretized at the Chebyshev points as
\begin{equation}
    \left\|q_w^b\left(\tau_i\right)\right\|=1 \quad i=0,1, \cdots, N_q \label{eq23}
\end{equation}
Substituting the Chebyshev approximations (\ref{eq18}), (\ref{eq19}), (\ref{eq20}), (\ref{eq21}) and the discretized constraints (\ref{eq23}) into (\ref{eq11}), we can reformulate the estimation task for the time interval of interest as
\begin{equation}
    \min _{\mathbf{D}, \mathbf{K}, b_a, b_g, X_1^w, \cdots, X_S^w}\left(J_{x_{I, 0}}+J_a+J_g+J_c\right) \quad \text { s.t }\left\|q_w^b\left(\tau_k\right)\right\|=1 \quad k=0,1, \cdots, N_q \label{eq24}
\end{equation}
where the estimation parameters include the Chebyshev coefficient, the accelerometer/gyroscope biases and the 3D feature points.
Equation (\ref{eq24}) is a constrained nonlinear squares problem, which can be
transformed to an unconstrained nonlinear least squares by the augmented Lagrangian method and then solved by the Levenberg-Marquardt algorithm \citep{bib30}. Once the coefficients are determined, the continuous poses as a function of time will be finally acquired by (\ref{eq18}), (\ref{eq19}) and (\ref{eq20}).

In summary, the approximations involved in the proposed batch estimators include the pose approximation by the Chebyshev polynomial in (\ref{eq18}), (\ref{eq19}) and (\ref{eq20}), the Clenshaw-Curtis quadrature in (\ref{eq21}) and the EFH interpolation of angular velocity and specific force. With the increased order of Chebyshev polynomial, the errors incurred by Chebyshev approximation and Clenshaw-Curtis quadrature will decay towards zero \citep{bib25}. Additionally, the EFH interpolation is also likely very close to optimality for time-equispaced samples \citep{bib31}. In this regard, the proposed state estimator by polynomial optimization is nearly optimal in the least squares sense \citep{bib33}.

\section{Simulation and experimentation} \label{sec3}
\subsection{Simulation results}
To assess the accuracy of the algorithm, two types of vehicular motion were designed for the simulation tests: a circular trajectory and a straight-line trajectory. The circular trajectory is generated in the world frame without considering the Earth's rotation; the straight-line trajectory is generated in the local navigation frame and then transformed into the Earth frame, taking into account the Earth's rotation. It is noted that the inclusion of Earth's rotation in this specific test was for the sake of diversity in simulation conditions and is not a requirement for the typical use-case scenarios of the proposed method.
\subsubsection{Circular trajectory}
\begin{figure}[htbp]
	\begin{minipage}{0.49\linewidth}
		\includegraphics[width=\linewidth]{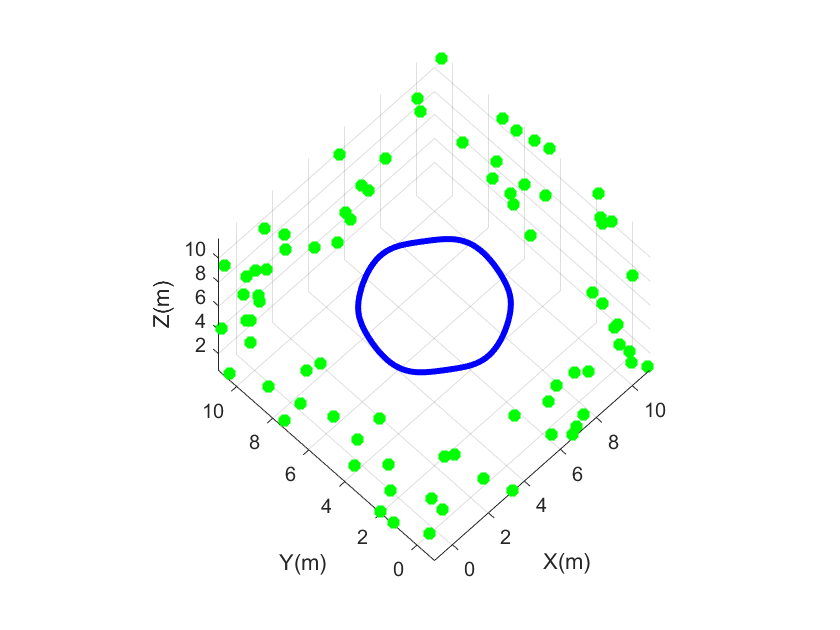}
		\caption{Circular trajectory.}
		\label{fig1}
	\end{minipage}
        \begin{minipage}{0.49\linewidth}
		\includegraphics[width=\linewidth]{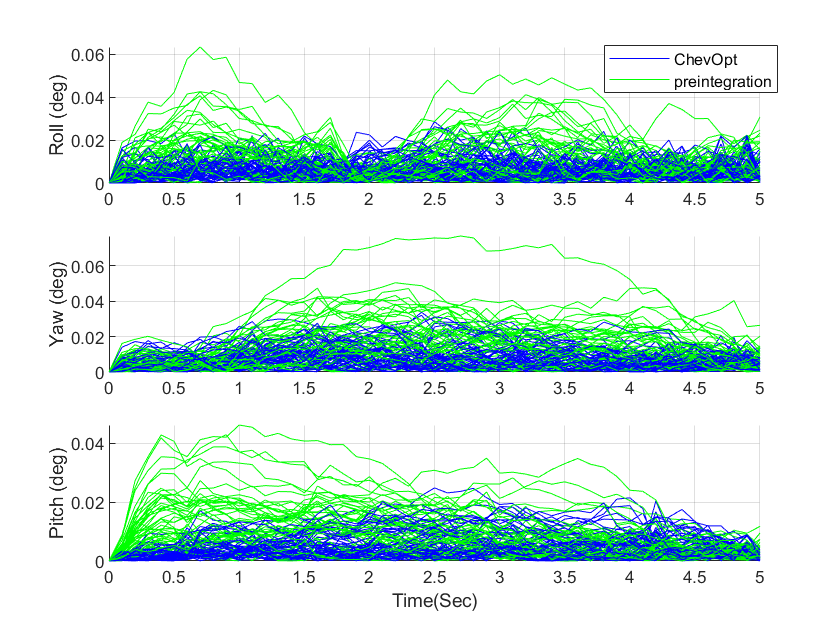}
		\caption{Attitude errors of Chebyshev polynomial optimization and preintegration across 50 Monte-Carlo runs.}
		\label{fig2}
	\end{minipage}
\end{figure}

As shown in Fig. \ref{fig1}, the circular trajectory has a radius of 3 m, with a slight sinusoidal motion in the vertical direction. Visual feature points are randomly distributed on the walls in the simulated space. To simulate real feature tracking, we  corrupt the feature measurements with isotropic Gaussian noise with standard deviation $\sigma_{px}= 1$ pixel. The camera has a focal length of 460 pixels and runs at a rate of 10 Hz. The gyroscope bias and accelerometer bias are set to 
$[0.3 \ -0.2 \ -0.5]^T \ (\mathrm{deg/s})$ and $[0.2 \ 0.1 \ -0.2]^T \ (\mathrm{m} / \mathrm{s}^2)$ , respectively. The root power spectral density of the gyroscope measurement noise of each axis is set to $1\ ^{\circ} / \sqrt{h}$, and the root power spectral density of the accelerometer measurement noise of each axis is set to $0.01 \ \mathrm{m} / \mathrm{s}^{1.5}$, with an IMU sampling frequency of 100 Hz. The total duration of the generated data is 5 seconds.

\begin{figure}[htbp]
	\begin{minipage}{0.49\linewidth}
		\includegraphics[width=\linewidth]{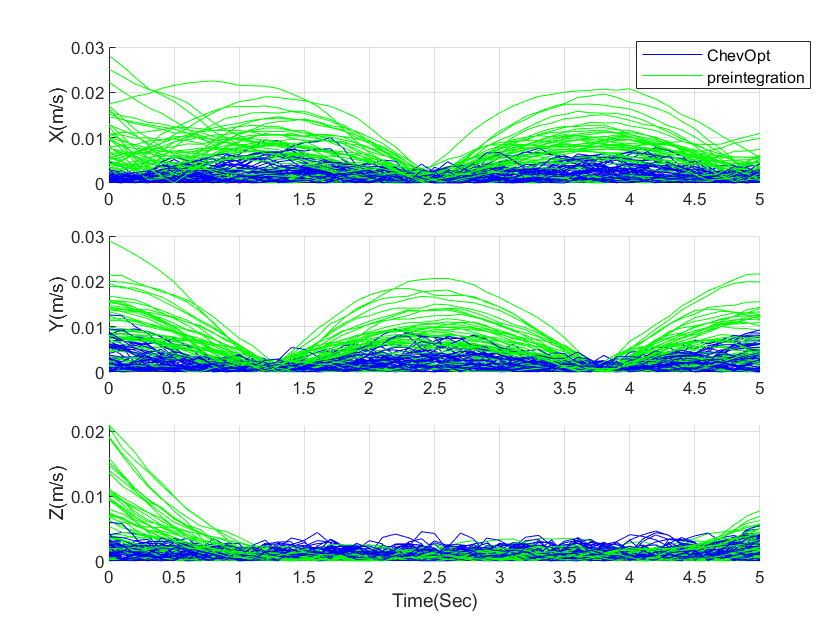}
		\caption{Velocity errors of Chebyshev polynomial optimization and preintegration across 50 Monte-Carlo runs.}
		\label{fig3}
	\end{minipage}
        \begin{minipage}{0.49\linewidth}
		\includegraphics[width=\linewidth]{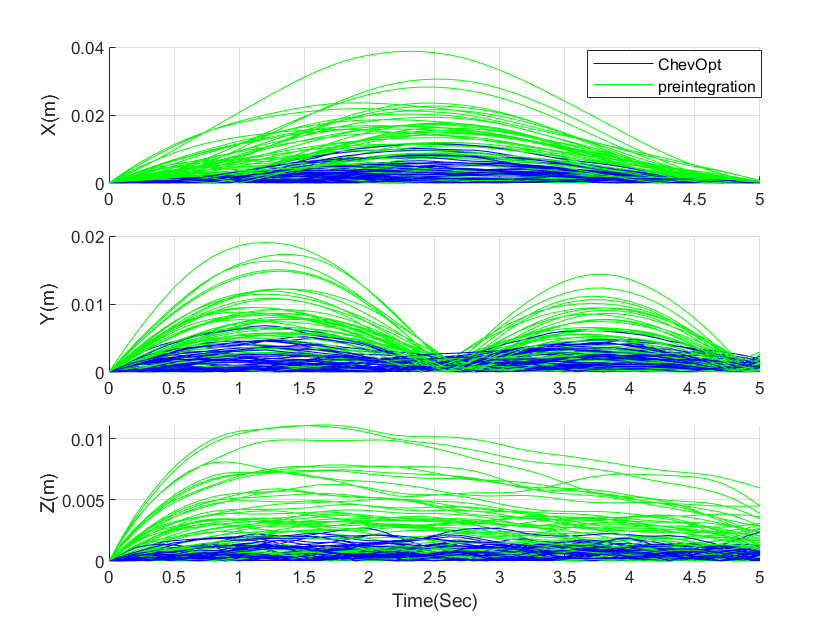}
		\caption{Position errors of Chebyshev polynomial optimization and preintegration across 50 Monte-Carlo runs.}
		\label{fig4}
	\end{minipage}
\end{figure}

\begin{figure}[htbp]
	\begin{minipage}{0.49\linewidth}
		\includegraphics[width=\linewidth]{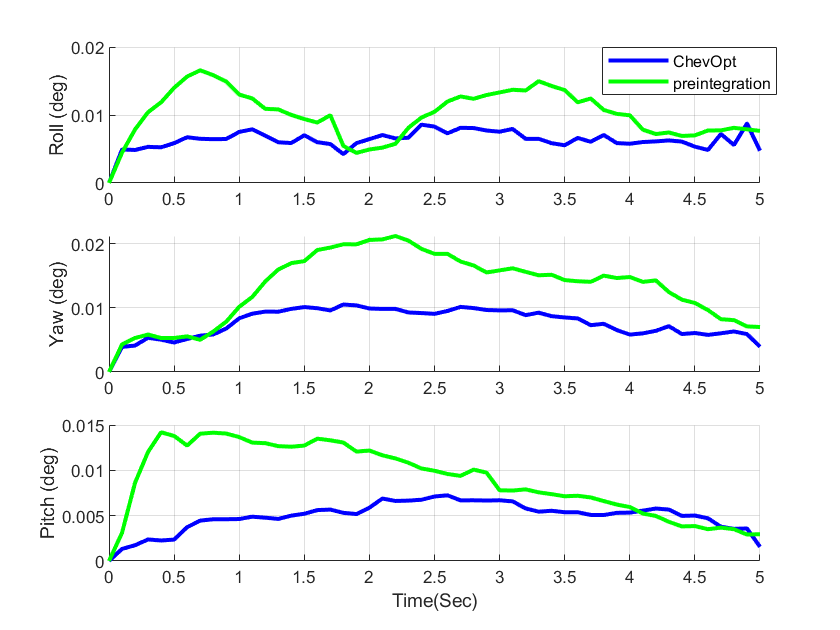}
		\caption{Average attitude errors across 50 Monte-Carlo runs for Chebyshev polynomial optimization and preintegration.}
		\label{fig5}
	\end{minipage}
        \begin{minipage}{0.49\linewidth}
		\includegraphics[width=\linewidth]{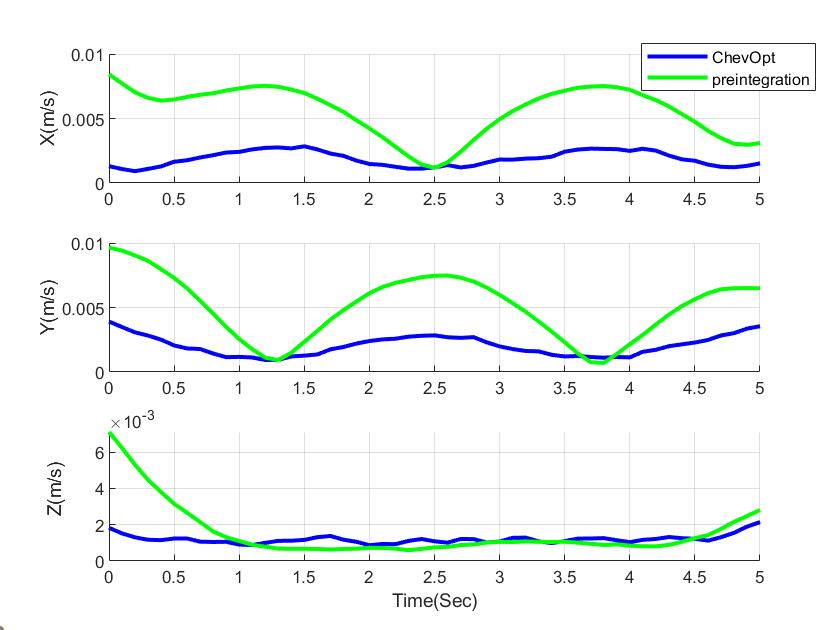}
		\caption{Average velocity errors across 50 Monte-Carlo runs for Chebyshev polynomial optimization and preintegration.}
		\label{fig6}
	\end{minipage}
\end{figure}

\begin{figure}[htbp]
	\begin{minipage}{0.49\linewidth}
		\includegraphics[width=\linewidth]{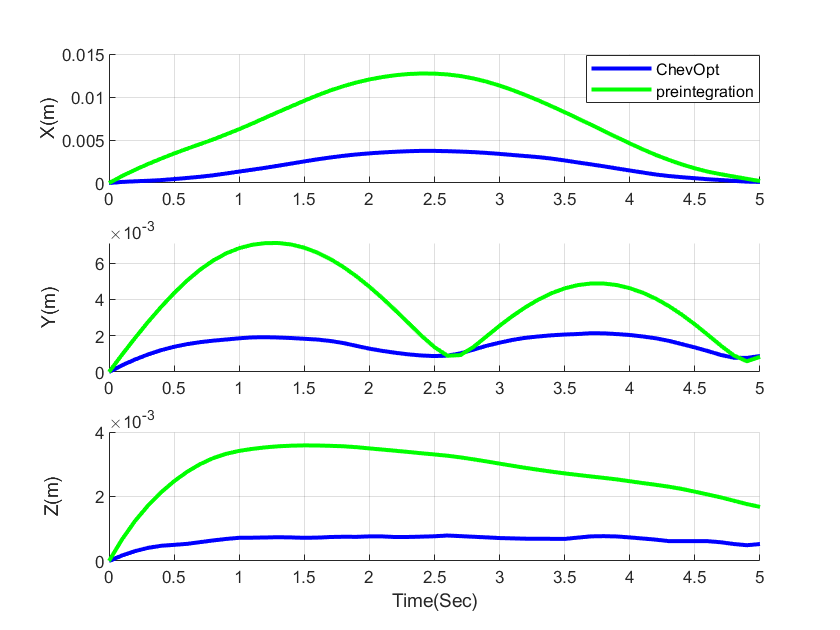}
		\caption{Average position errors across 50 Monte-Carlo runs for Chebyshev polynomial optimization and preintegration.}
		\label{fig7}
	\end{minipage}
        \begin{minipage}{0.49\linewidth}
		\includegraphics[width=\linewidth]{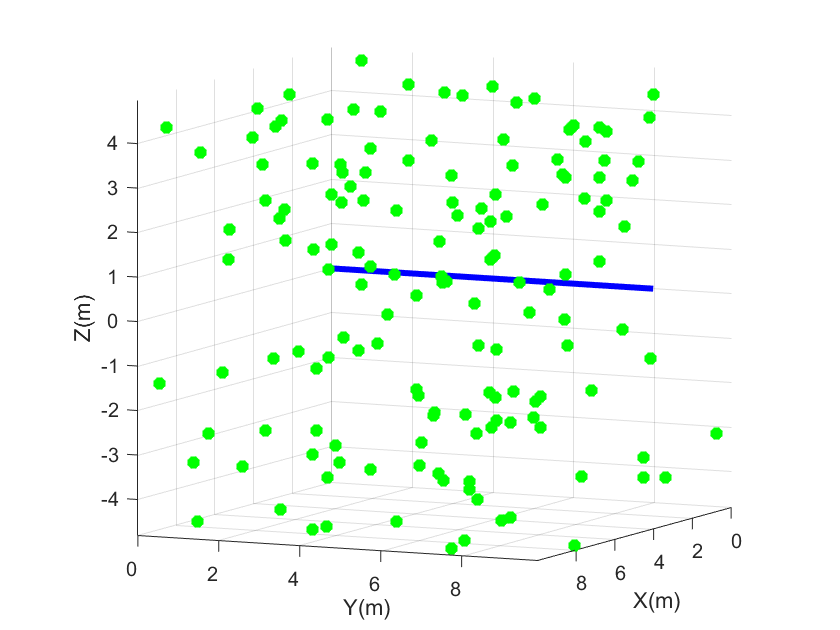}
		\caption{Straight-line trajectory.}
		\label{fig8}
	\end{minipage}
\end{figure}

The initial gyroscope and accelerometer biases are set to zeros and the statistics of the random noises are assumed to be known. The initial state within the interval is obtained through inertial navigation integration, and the initial coordinates of the 3D feature points are obtained by linear triangulation. The Chebyshev polynomial orders
of both attitude and velocity are set to $N_q=N_v=60$. Figs. \ref{fig2}-\ref{fig4} respectively plot the attitude, velocity, and position errors across 50 Monte Carlo runs. For brevity, Figs. \ref{fig5}-\ref{fig7} display the average errors in attitude, velocity, and position, respectively, which is computed as 
\begin{equation}
    \varepsilon_i(k)=\frac{1}{H_m} \sum_{h=1}^{H_m}\left\|x_{(i), k}^h-\hat{x}_{(i), k}^h\right\| \label{eq25}
\end{equation}
where $H_m$ denotes the number of Monte Carlo runs, $k$ indicates the time, and $i$ represents different state components. It is evident that Chebyshev polynomial optimization achieves higher estimation precision.
Table \ref{tab1} lists the accumulative RMSEs of each estimated state for the circular trajectory, which are computed as
\begin{equation}
    A R M S E_q=\sqrt{\frac{1}{H_m \cdot Z} \sum_{h=1}^{H_m} \sum_{k=1}^Z\left\|\log \left(q_w^{b_k{ }^*}(h) \circ \hat{q}_w^{b_k}(h)\right)\right\|^2} \label{eq26}
\end{equation}
\begin{equation}
     A R M S E_v=\sqrt{\frac{1}{H_m \cdot Z} \sum_{h=1}^{H_m} \sum_{k=1}^Z\left\|v_{w b_k}^w(h)-\hat{v}_{w b_k}^w(h)\right\|^2} 
    \label{eq27}
\end{equation}
\begin{equation}
    A R M S E_p=\sqrt{\frac{1}{H_m \cdot Z} \sum_{h=1}^{H_m} \sum_{k=1}^Z\left\|p_{w b_k}^w(h)-\hat{p}_{w b_k}^w(h)\right\|^2} 
\label{eq28}
\end{equation}
where $Z$ is the number of states corresponding to the time interval. In the circular trajectory simulation, the attitude accumulative RMSE of the Chebyshev polynomial optimization is approximately 47\% lower than preintegration, the velocity accumulative RMSE is about 58\% lower, and the position accumulative RMSE is around 65\% lower.
\begin{table}[htbp]
\caption{ARMSEs of each estimated state under circular trajectory.}\label{tab1}
\begin{tabular}{@{}llll@{}}
\toprule
  & \textbf{Attitude (deg)} & \textbf{Velocity (m/s) }& \textbf{Position (m)}\\
\midrule
\textbf{Chebyshev}& \textbf{0.0136} & \textbf{0.0038} & \textbf{0.0034} \\
\textbf{Preintegration} & 0.0255 & 0.0090 & 0.0098 \\
\botrule
\end{tabular}
\end{table}

\subsubsection{Straight-line trajectory}
 Considering the convenient gravity model in the local navigation frame (n-frame, with the coordinate order: North-Up-East), the straight-line trajectory is generated in the local navigation frame and then transformed to the Earth frame \citep{bib26}. Suppose a vehicle carrying a visual-inertial navigation system starts from 0 longitude, 0 latitude, 0 altitude and travels eastward with an initial velocity of 0 m/s, where the velocity change follows $\dot{v}_{e b}^n=[0 \ 0 \ a \sin (w t)]^T$, with magnitude $a=2 \mathrm{~m} / \mathrm{s}^2$ and angular frequency $w=0.4 \pi \ \mathrm{rad} / \mathrm{s}$. The body attitude is assumed to undergo a classical coning motion described by the attitude quaternion
\begin{equation}
    q_n^b=\cos (\alpha / 2)+\sin (\alpha / 2)\begin{bmatrix}
0& \cos (\zeta t) & \sin (\zeta t)
\end{bmatrix}^T \label{eq29}
\end{equation}
where $\zeta=0.5 \pi \ \mathrm{rad} / \mathrm{s}$ is the coning frequency, and $\alpha=30 \ \mathrm{deg}$ is the coning angle. Then, the true velocity and position are readily given by
\begin{equation}
v_{e b}^n=v_{e b}^n(0)+\int_0^t \dot{v}_{e b}^n d t=\left[\begin{array}{lll}
0 & 0 & v_0-(a \cos (w t)-a) / w\end{array}\right]^T \label{30}
\end{equation}
\begin{equation}
p^n=\begin{bmatrix}
 v_0 t-(a \sin (w t)-a w t) / w^2& 0 & 0
\end{bmatrix}^T / R_E
\label{31}
\end{equation}
where the position $p^n=\left[\lambda \ L \ h\right]^T$ is expressed in terms of longitude $\lambda$, latitude $L$, and altitude $h$, and $R_E$ is the transverse radius of curvature. 3D feature points are randomly distributed in the simulation space. The IMU sampling frequency is set to 100 Hz, and the camera sampling frequency is 10 Hz. The gyroscope bias and accelerometer bias are set to $\left[0.3 \ 0.2 \ -0.5\right]^T\ (\mathrm{deg/s})$ and $\left[0.2 \ -0.2\ -0.1\right]^T\left(\mathrm{m} / \mathrm{s}^2\right)$, respectively. The root power spectral density of the gyroscope measurement noise of each axis is set to $1\ ^{\circ} / \sqrt{h}$, the root power spectral density of the accelerometer measurement noise of each axis is set to $0.01 \ \mathrm{m} / \mathrm{s}^{1.5}$ and the standard deviation of the feature measurements noise is set to 0.5 pixels. The data length is 5 seconds, and the generated trajectory is shown in Fig. \ref{fig8}.

\begin{figure}[htbp]
	\begin{minipage}{0.49\linewidth}
		\includegraphics[width=\linewidth]{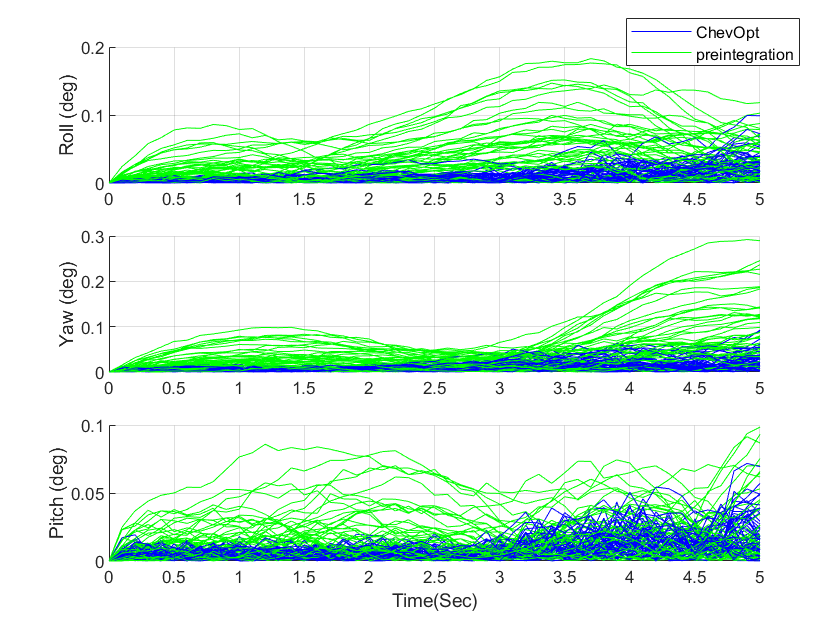}
		\caption{Attitude errors of Chebyshev polynomial optimization and preintegration across 50 Monte-Carlo runs.}
		\label{fig9}
	\end{minipage}
        \begin{minipage}{0.49\linewidth}
		\includegraphics[width=\linewidth]{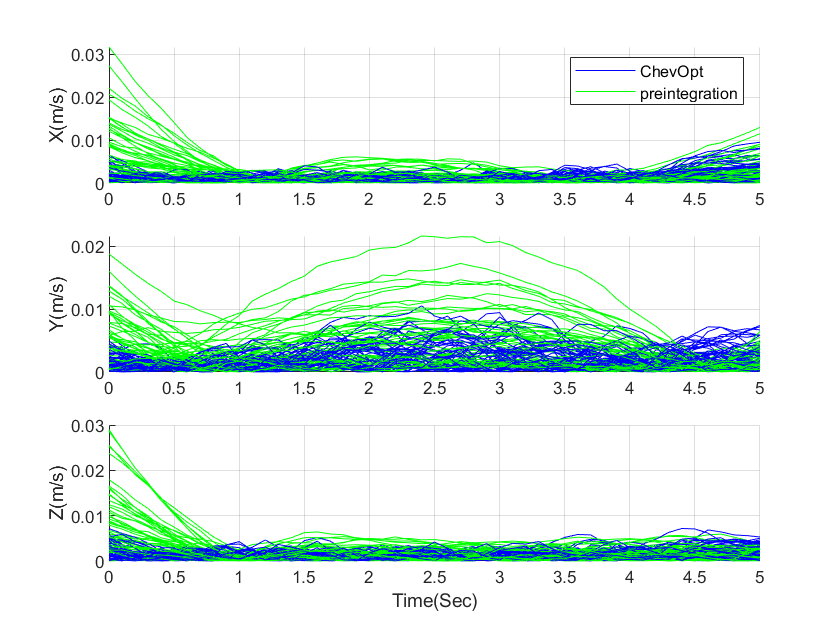}
		\caption{Velocity errors of Chebyshev polynomial optimization and preintegration across 50 Monte-Carlo runs.}
		\label{fig10}
	\end{minipage}
\end{figure}

\begin{figure}[htbp]
	\begin{minipage}{0.49\linewidth}
		\includegraphics[width=\linewidth]{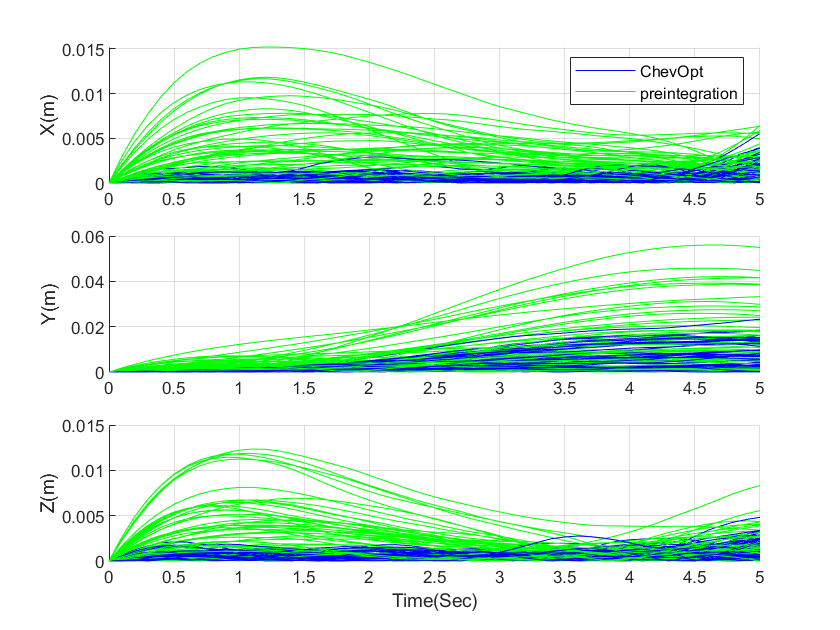}
        \caption{Position errors of Chebyshev polynomial optimization and preintegration across 50 Monte-Carlo runs.}
		\label{fig11}
	\end{minipage}
        \begin{minipage}{0.49\linewidth}
		\includegraphics[width=\linewidth]{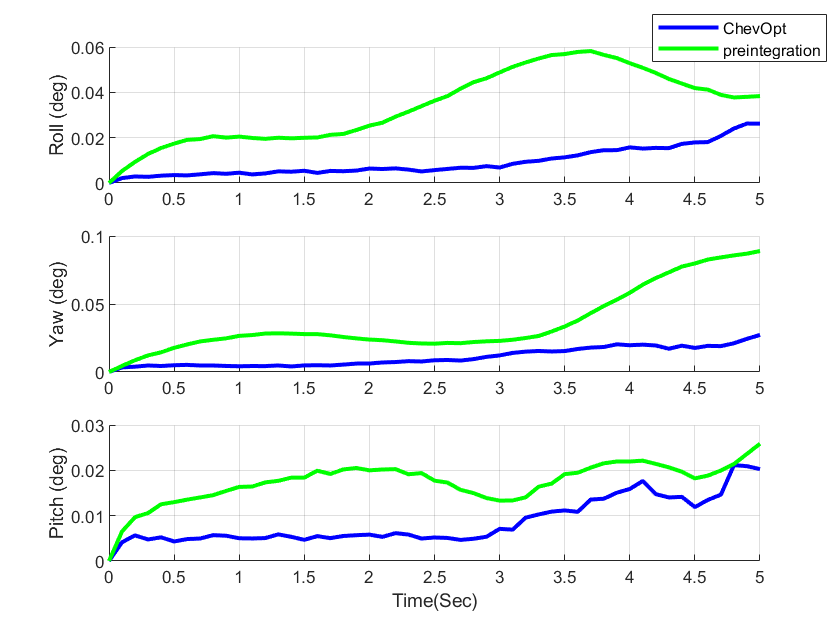}
        \caption{Average attitude errors across 50 Monte-Carlo runs for Chebyshev polynomial optimization and preintegration.}
		\label{fig12}
	\end{minipage}
\end{figure}

\begin{figure}[htbp]
	\begin{minipage}{0.49\linewidth}
		\includegraphics[width=\linewidth]{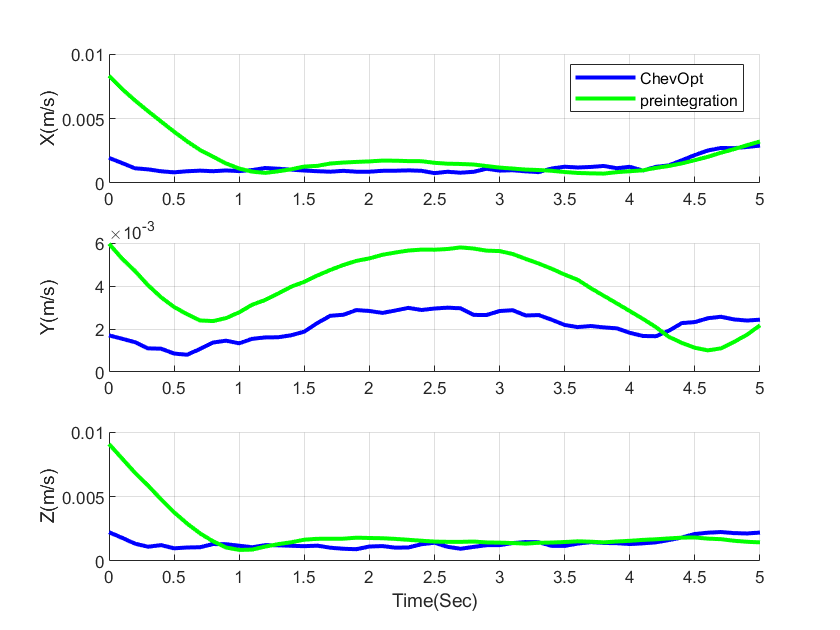}
        \caption{Average velocity errors across 50 Monte-Carlo runs for Chebyshev polynomial optimization and preintegration.}
		\label{fig13}
	\end{minipage}
        \begin{minipage}{0.49\linewidth}
		\includegraphics[width=\linewidth]{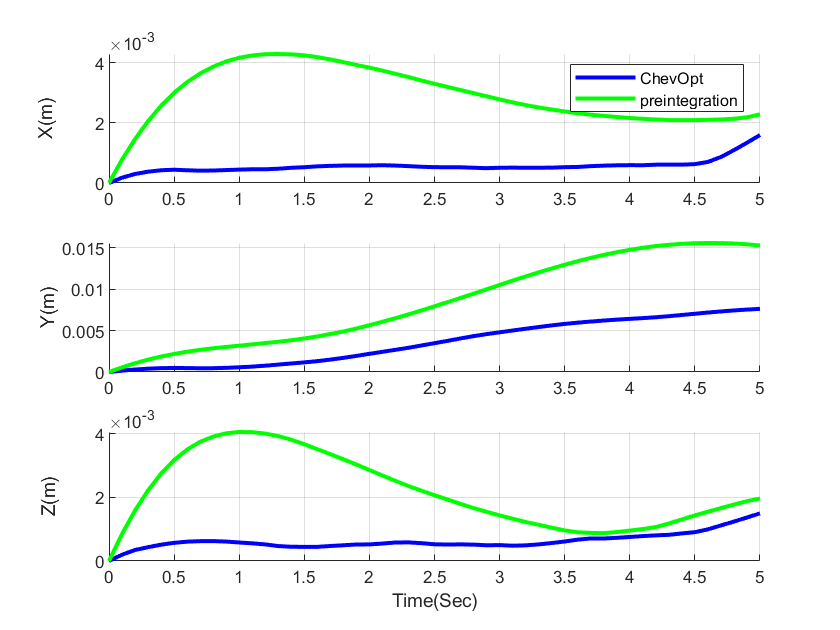}
		\caption{Average position errors across 50 Monte-Carlo runs for Chebyshev polynomial optimization and preintegration.}
		\label{fig14}
	\end{minipage}
\end{figure}

Figs. \ref{fig9}-\ref{fig11} respectively plot the attitude, velocity, and position errors across 50 Monte Carlo runs, while Figs. \ref{fig12}-\ref{fig14} display the average errors. It is observable that, compared to the preintegration method, the Chebyshev polynomial optimization demonstrates better accuracy. This aligns with the conclusion drawn from the circular trajectory simulation. Table \ref{tab2}  lists the accumulative RMSEs of each estimated state for the straight-line trajectory, with the Chebyshev polynomial optimization showing a reduction of about 68\% in attitude accumulative RMSE, 49\% in velocity, and 59\% in position relative to preintegration.
\begin{table}[htbp]
\caption{ARMSEs of each estimated state under straight-line trajectory.}\label{tab2}%
\begin{tabular}{@{}llll@{}}
\toprule
  & \textbf{Attitude (deg)} & \textbf{Velocity (m/s)} & \textbf{Position (m)}\\
\midrule
\textbf{Chebyshev} & \textbf{0.0248} & \textbf{0.0037} & \textbf{0.0057} \\
\textbf{Preintegration} & 0.0773 & 0.0073 & 0.0140 \\
\botrule
\end{tabular}
\end{table}

\subsection{Experiment results}
\begin{figure}[htbp]
        \centering
	\begin{minipage}{0.24\linewidth}
		\includegraphics[width=\linewidth]{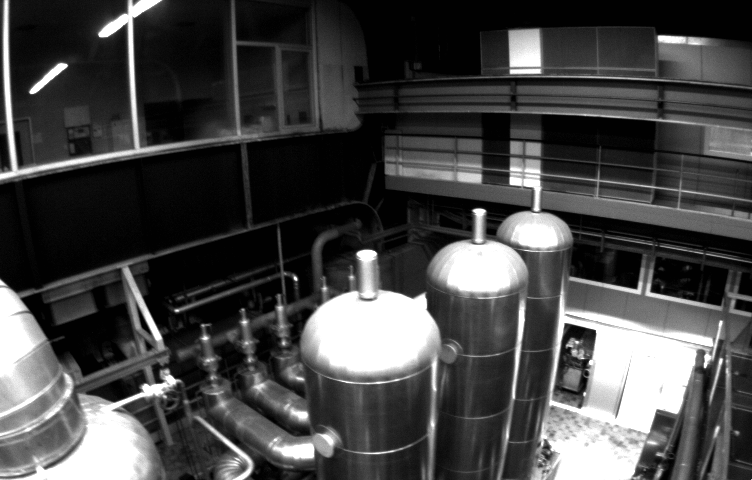}
	\end{minipage}
        \begin{minipage}{0.24\linewidth}
		\includegraphics[width=\linewidth]{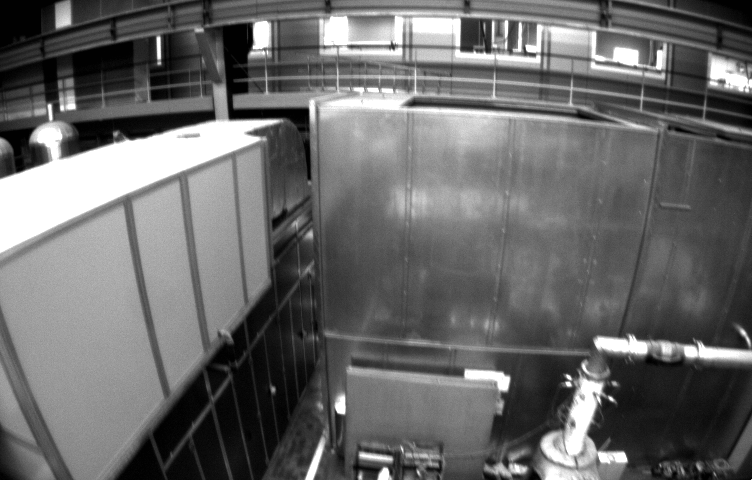}
	\end{minipage}
        \begin{minipage}{0.24\linewidth}
		\includegraphics[width=\linewidth]{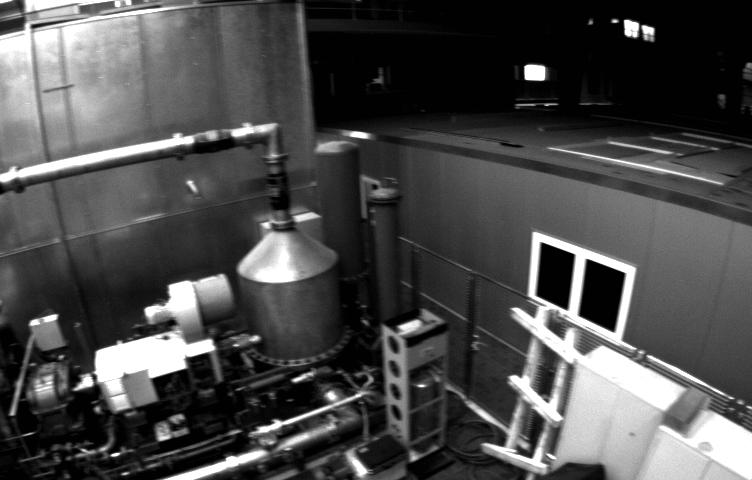}
	\end{minipage}
        \begin{minipage}{0.24\linewidth}
		\includegraphics[width=\linewidth]{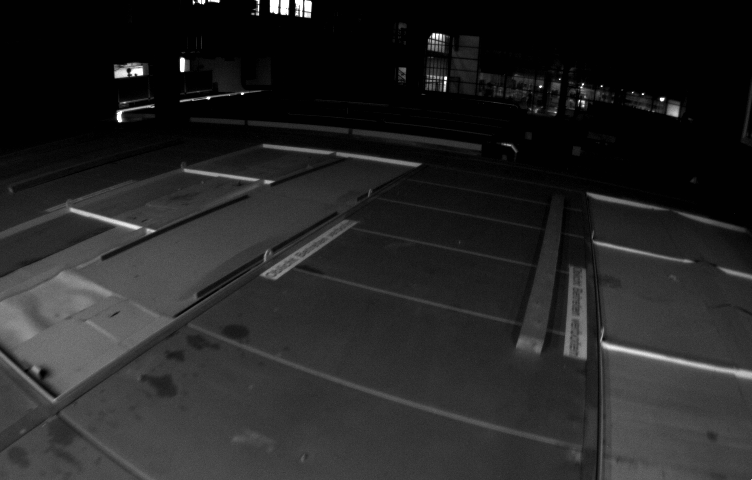}
	\end{minipage}       
 \caption{Multiple scene samplings derived from the machine hall sequences\citep{bib32}.}\label{fig15}
\end{figure}

\begin{figure}[htbp]
        \centering
	\begin{minipage}{0.24\linewidth}
		\includegraphics[width=\linewidth]{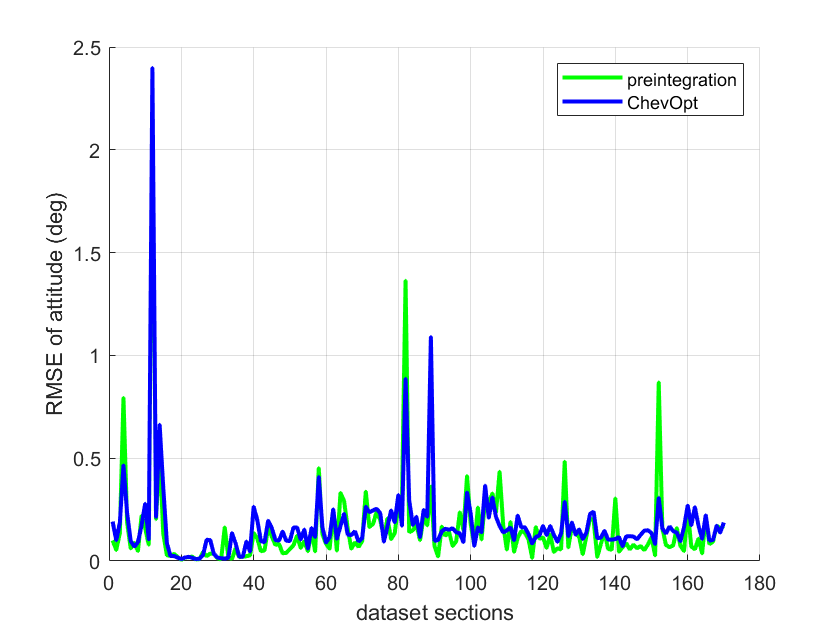}
		\subcaption{attitude}
		\label{fig16a}
	\end{minipage}
        \begin{minipage}{0.24\linewidth}
		\includegraphics[width=\linewidth]{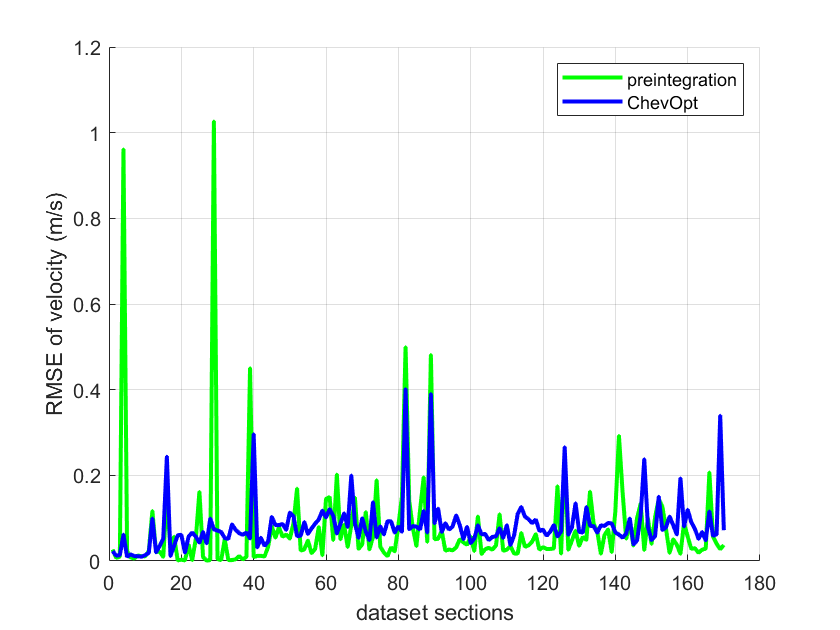}
		\subcaption{velocity}
		\label{fig16b}
	\end{minipage}
        \begin{minipage}{0.24\linewidth}
		\includegraphics[width=\linewidth]{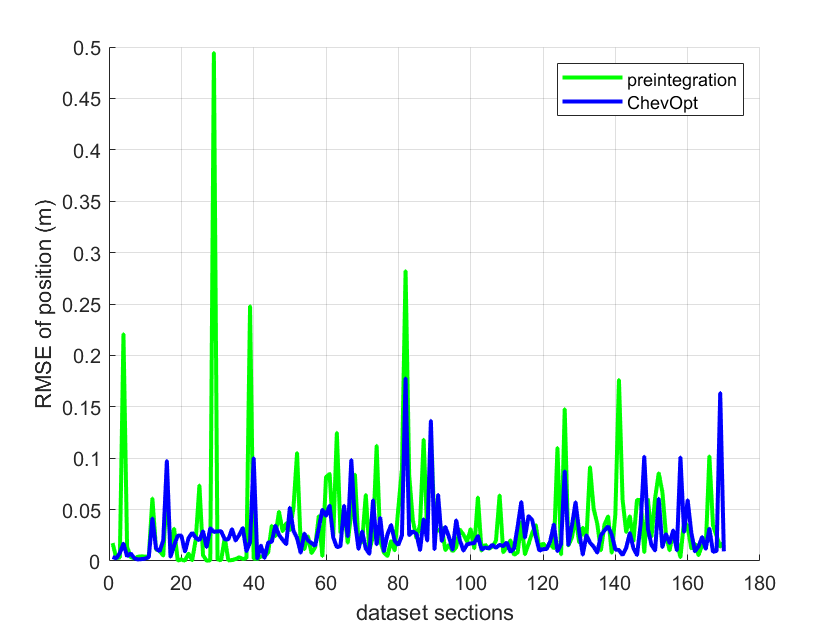}
		\subcaption{position}
		\label{fig16c}
	\end{minipage}
        \begin{minipage}{0.24\linewidth}
		\includegraphics[width=\linewidth]{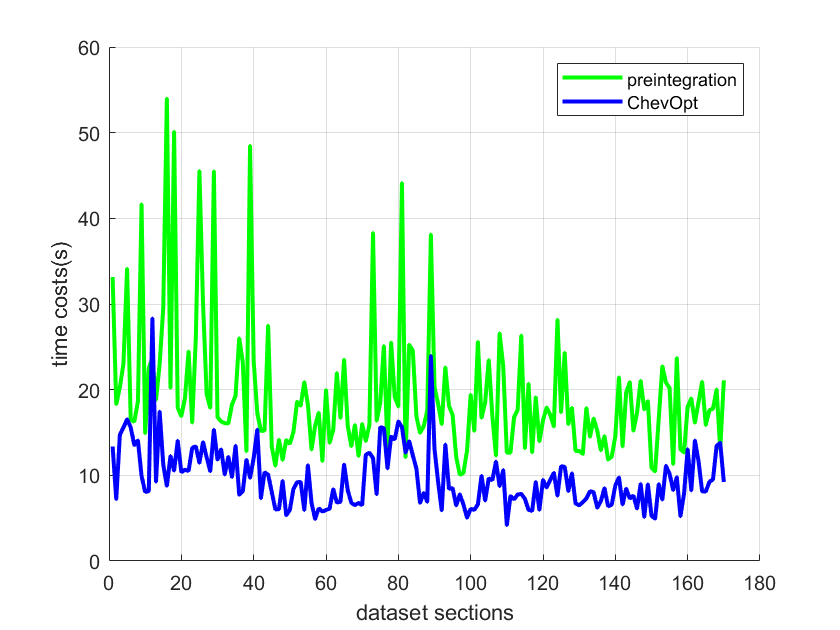}
		\subcaption{time cost}
		\label{fig16d}
	\end{minipage}
 \caption{RMSEs of estimated states and time cost for each segment of the MH\_01\_easy sequence: results from Chebyshev polynomial optimization and preintegration.}
         \label{fig16}
\end{figure}

\begin{figure}[htbp]
        \centering
	\begin{minipage}{0.24\linewidth}
		\includegraphics[width=\linewidth]{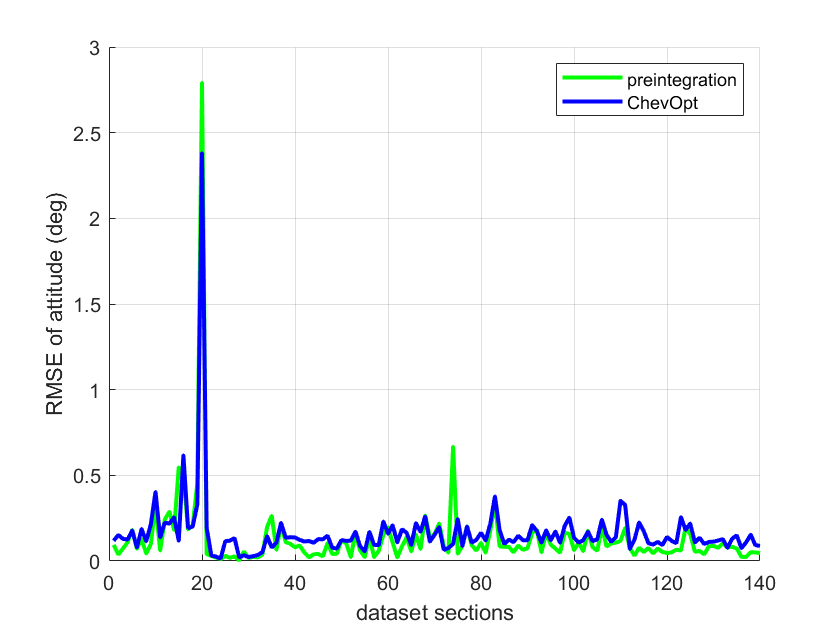}
		\subcaption{attitude}
		\label{fig17a}
	\end{minipage}
        \begin{minipage}{0.24\linewidth}
		\includegraphics[width=\linewidth]{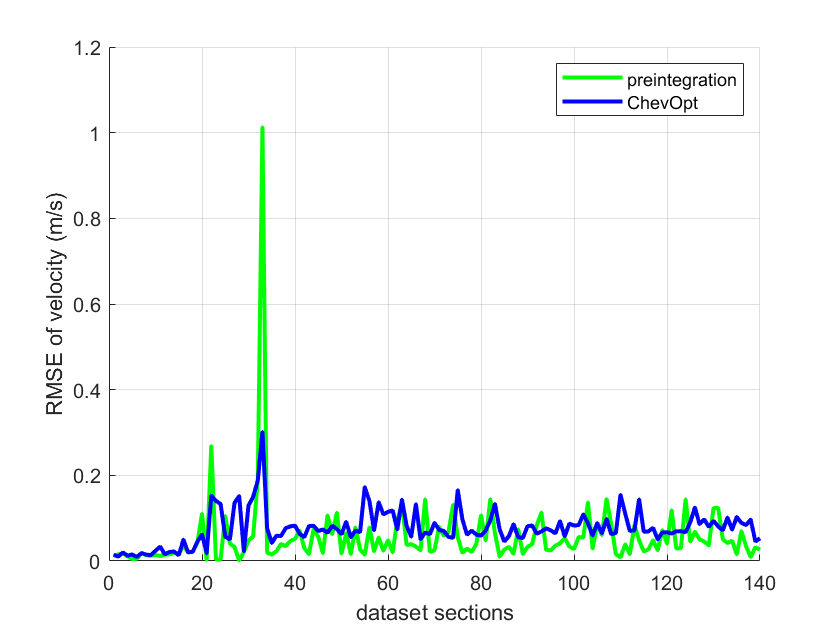}
		\subcaption{velocity}
		\label{fig17b}
	\end{minipage}
        \begin{minipage}{0.24\linewidth}
		\includegraphics[width=\linewidth]{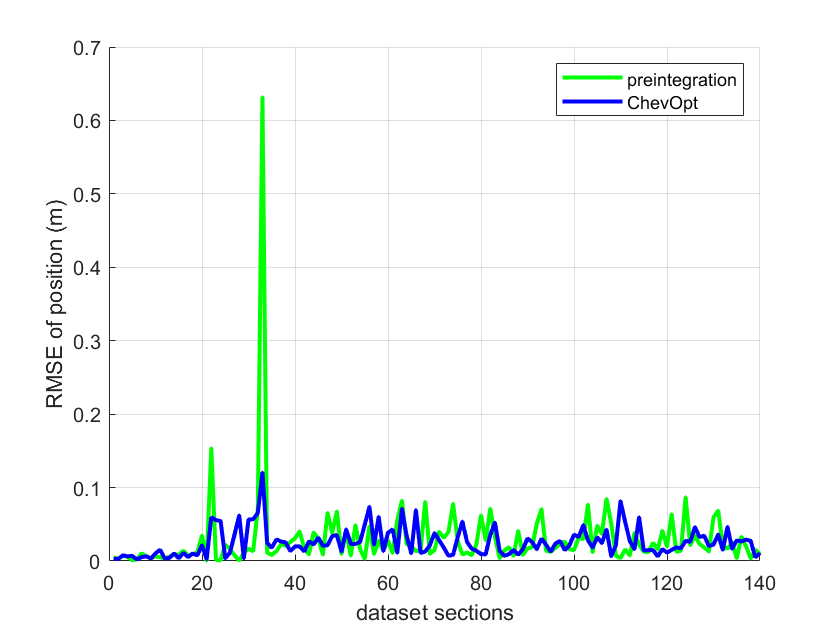}
		\subcaption{position}
		\label{fig17c}
	\end{minipage}
        \begin{minipage}{0.24\linewidth}
		\includegraphics[width=\linewidth]{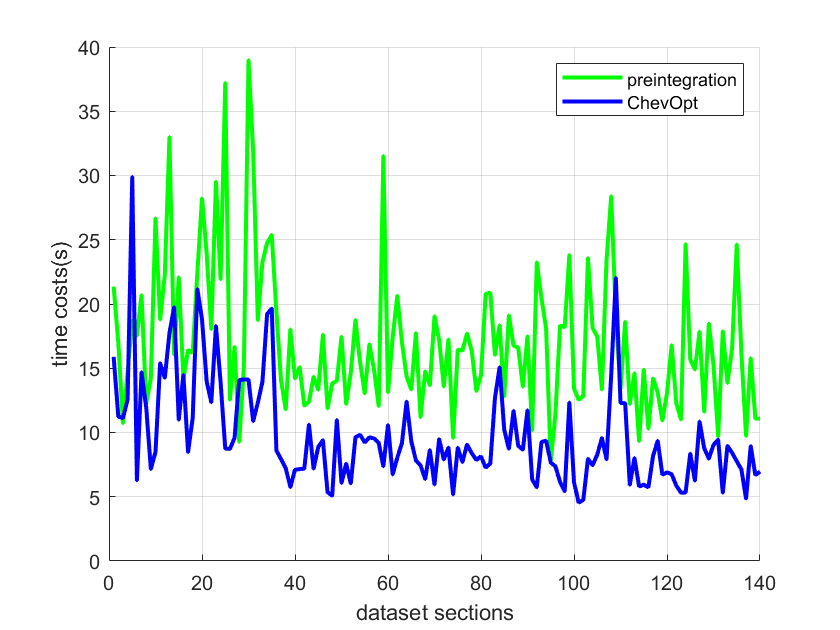}
		\subcaption{time cost}
		\label{fig17d}
	\end{minipage}
 \caption{RMSEs of estimated states and time cost for each segment of the MH\_02\_easy sequence: results from Chebyshev polynomial optimization and preintegration.}
         \label{fig17}
\end{figure}

\begin{figure}[htbp]
        \centering
	\begin{minipage}{0.24\linewidth}
		\includegraphics[width=\linewidth]{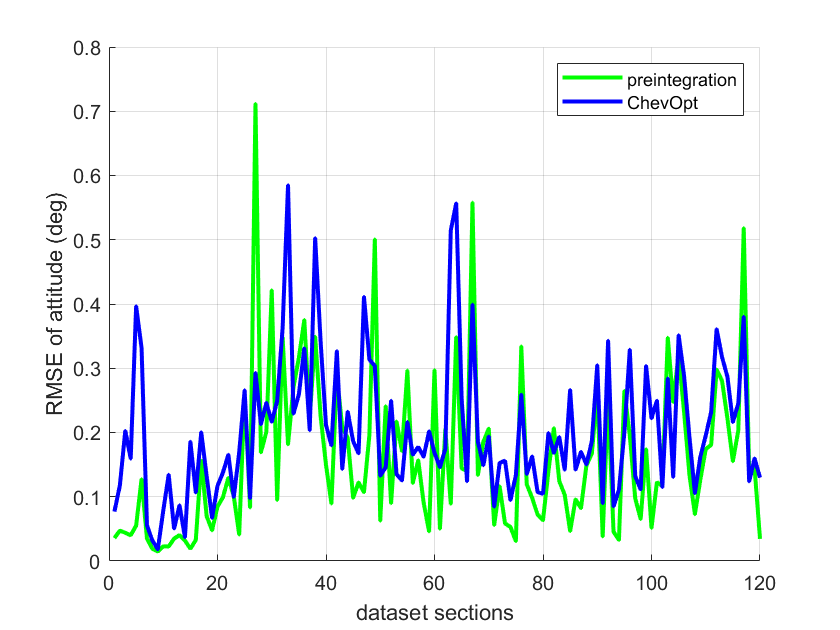}
		\subcaption{attitude}
		\label{fig18a}
	\end{minipage}
        \begin{minipage}{0.24\linewidth}
		\includegraphics[width=\linewidth]{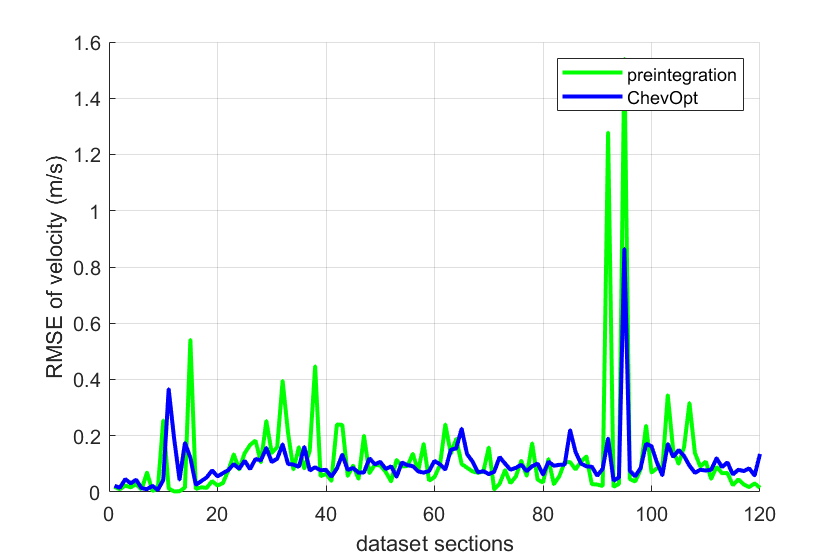}
		\subcaption{velocity}
		\label{fig18b}
	\end{minipage}
        \begin{minipage}{0.24\linewidth}
		\includegraphics[width=\linewidth]{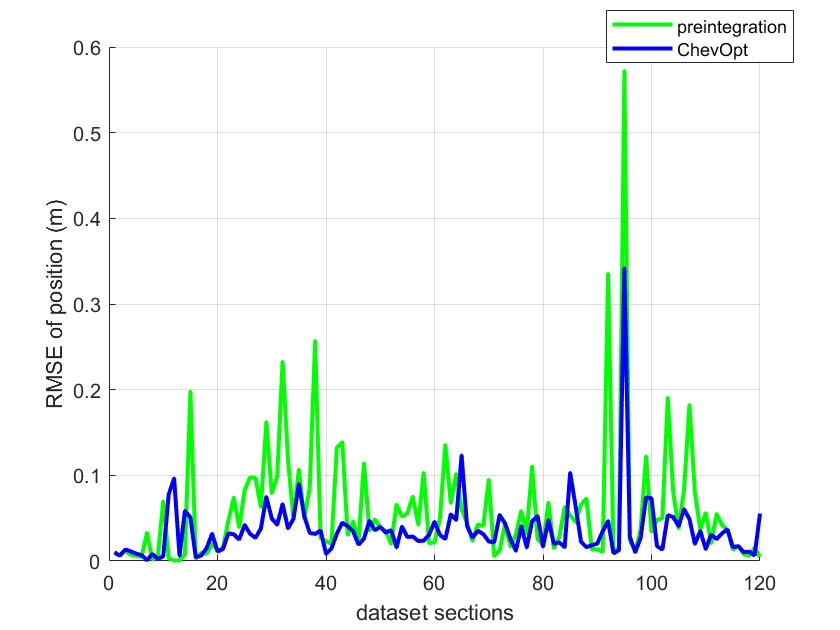}
		\subcaption{position}
		\label{fig18c}
	\end{minipage}
        \begin{minipage}{0.24\linewidth}
		\includegraphics[width=\linewidth]{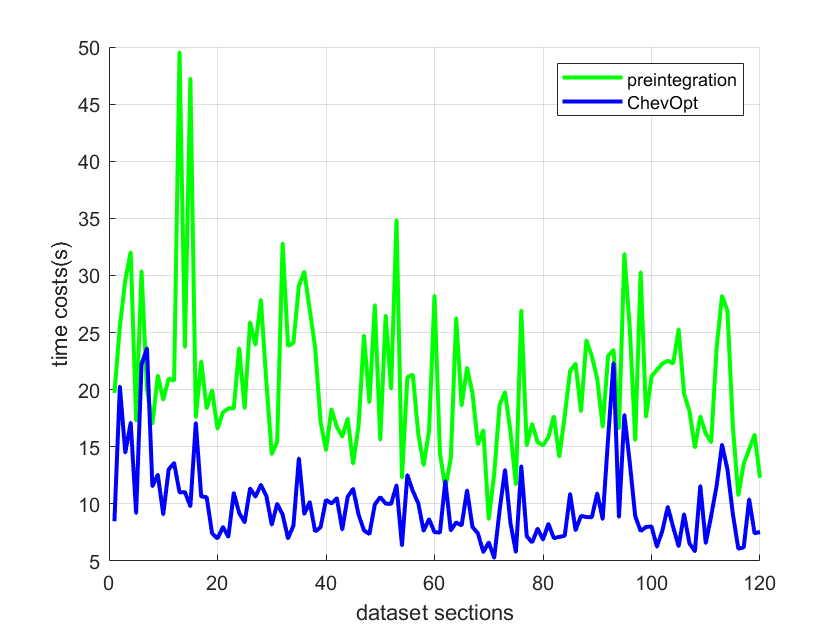}
		\subcaption{time cost}
		\label{fig18d}
	\end{minipage}
 \caption{RMSEs of estimated states and time cost for each segment of the MH\_03\_medium sequence: results from Chebyshev polynomial optimization and preintegration.}
         \label{fig18}
\end{figure}

\begin{figure}[htbp]
        \centering
	\begin{minipage}{0.24\linewidth}
		\includegraphics[width=\linewidth]{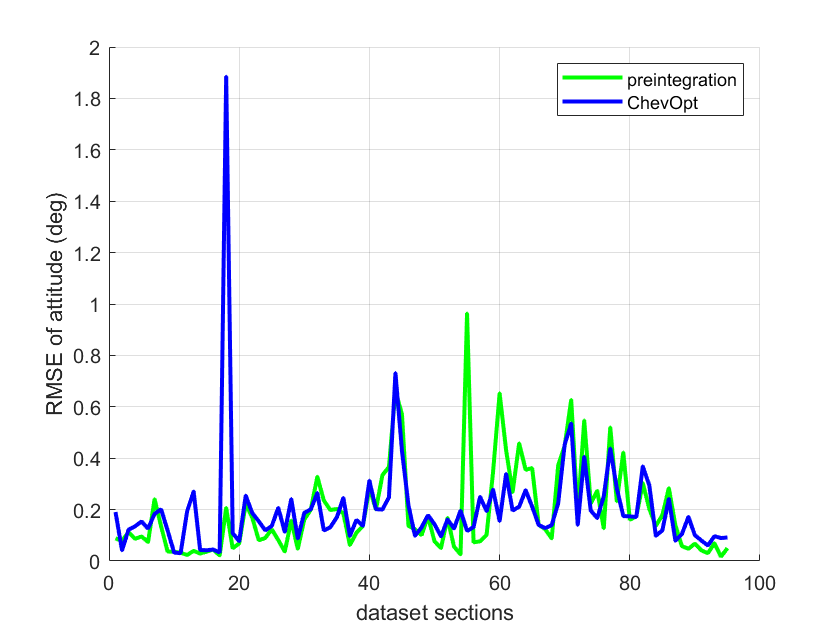}
		\subcaption{attitude}
		\label{fig19a}
	\end{minipage}
        \begin{minipage}{0.24\linewidth}
		\includegraphics[width=\linewidth]{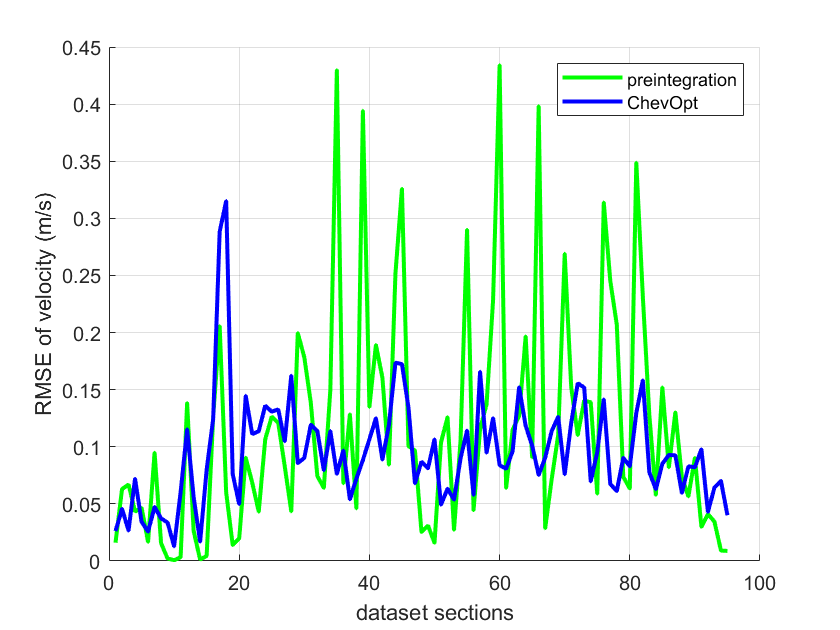}
		\subcaption{velocity}
		\label{fig19b}
	\end{minipage}
        \begin{minipage}{0.24\linewidth}
		\includegraphics[width=\linewidth]{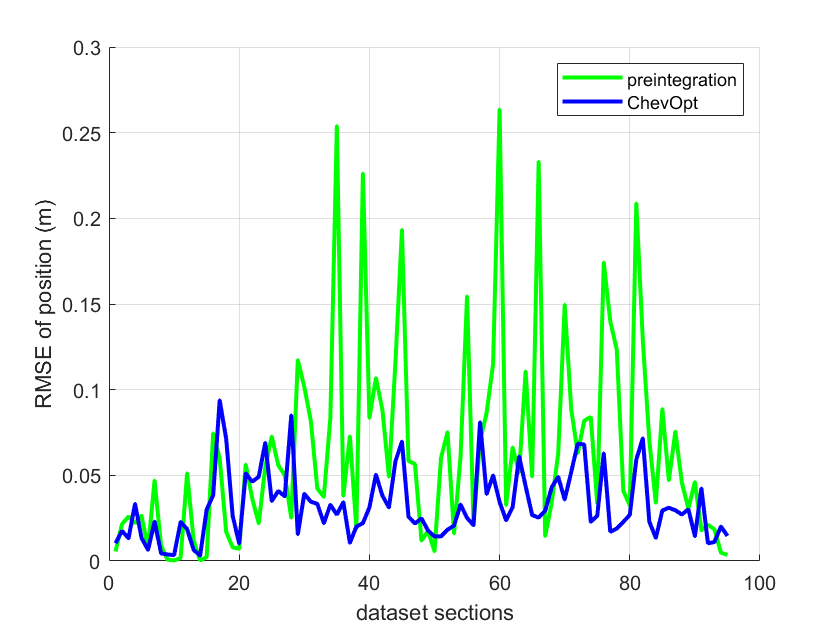}
		\subcaption{position}
		\label{fig19c}
	\end{minipage}
        \begin{minipage}{0.24\linewidth}
		\includegraphics[width=\linewidth]{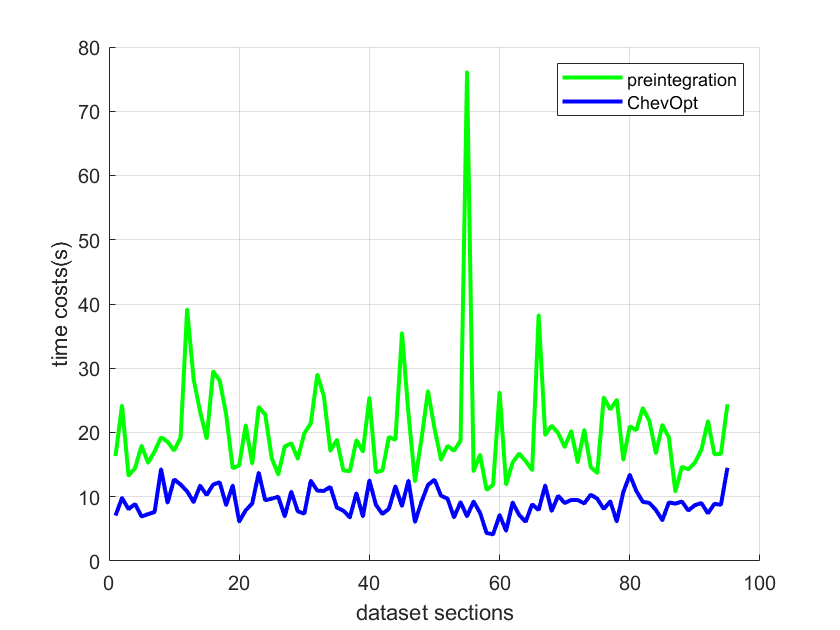}
		\subcaption{time cost}
		\label{fig19d}
	\end{minipage}
 \caption{RMSEs of estimated states and time cost for each segment of the MH\_04\_difficult sequence: results from Chebyshev polynomial optimization and preintegration.}
         \label{fig19}
\end{figure}

\begin{figure}[htbp]
        \centering
	\begin{minipage}{0.24\linewidth}
		\includegraphics[width=\linewidth]{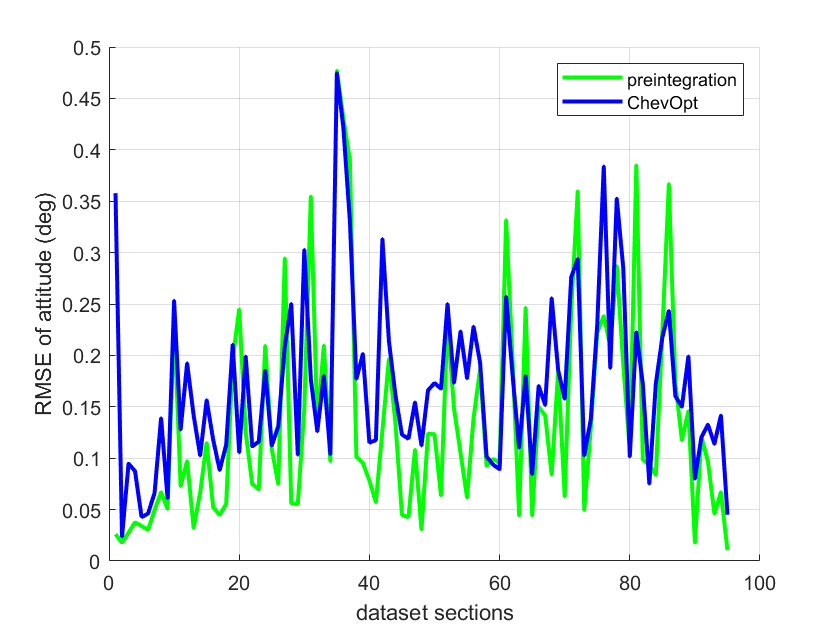}
		\subcaption{attitude}
		\label{fig20a}
	\end{minipage}
        \begin{minipage}{0.24\linewidth}
		\includegraphics[width=\linewidth]{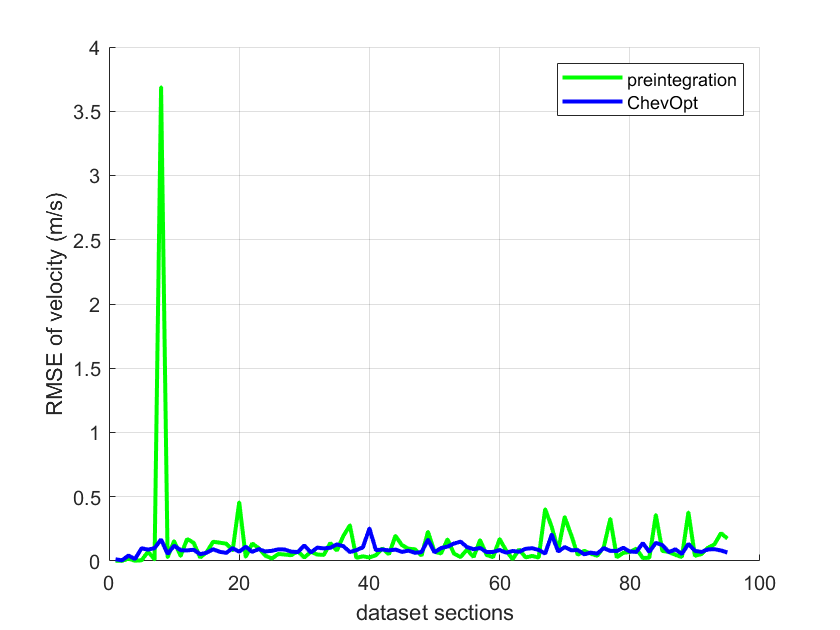}
		\subcaption{velocity}
		\label{fig20b}
	\end{minipage}
        \begin{minipage}{0.24\linewidth}
		\includegraphics[width=\linewidth]{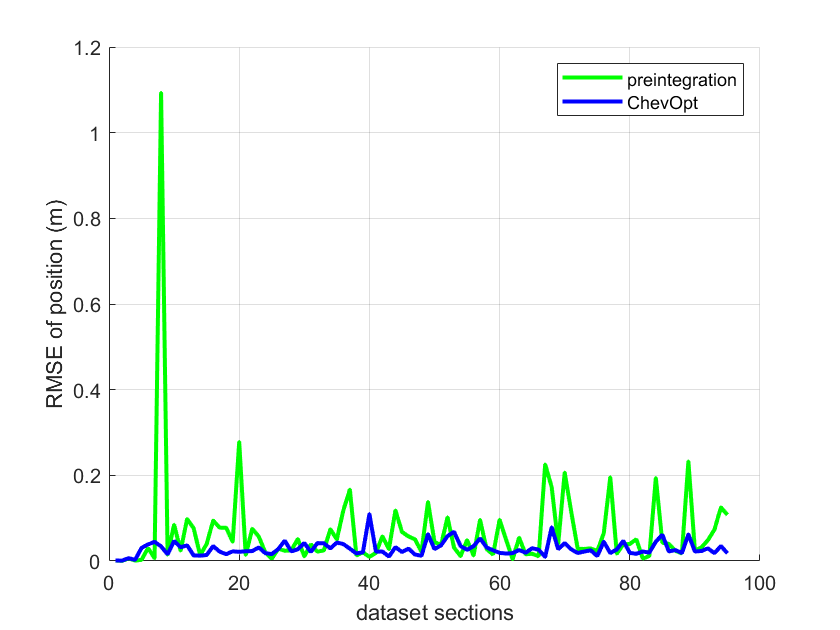}
		\subcaption{position}
		\label{fig20c}
	\end{minipage}
        \begin{minipage}{0.24\linewidth}
		\includegraphics[width=\linewidth]{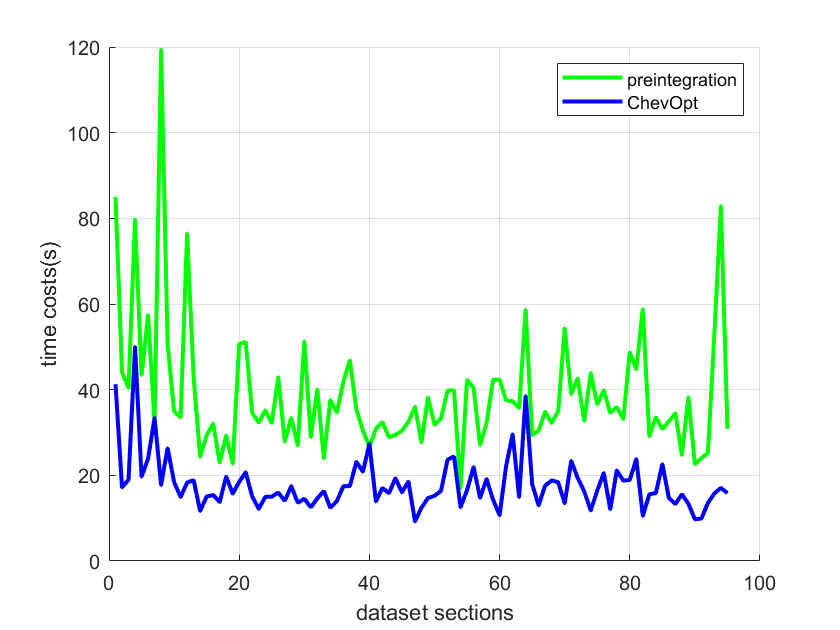}
		\subcaption{time cost}
		\label{fig20d}
	\end{minipage}
 \caption{RMSEs of estimated states and time cost for each segment of the MH\_05\_difficult sequence: results from Chebyshev polynomial optimization and preintegration.}
         \label{fig20}
\end{figure}

The experimental evaluation of the proposed algorithm was conducted utilizing the EuRoC MAV VI datasets \citep{bib32}, collected by a micro-aerial vehicle (MAV). The IMU sampling frequency is 200 Hz, and the stereo camera sampling frequency is 20 Hz. The intrinsics and extrinsics of the visual-inertial sensor unit are calibrated and timestamps are aligned. We performed segmented testing (each segment of 1 second) on all machine hall sequences in the dataset to demonstrate the algorithm's effectiveness, using only the left camera data for monocular measurements. As shown in Fig. \ref{fig15}, these sequences are collected in the machine hall, presenting a complex environment and motion. The Chebyshev polynomial orders are set to $N_q=N_v=16$. The interval for both the visual keyframes and the IMU preintegration is set to 0.1 seconds. The value of gravity is obtained through the initialization method in VINS-Mono \citep{bib9}. Figs. \ref{fig16}-\ref{fig20} plot the RMSEs of attitude, velocity, and position for each segment of every sequence, along with the corresponding time costs. It is shown that the Chebyshev polynomial optimization delivers more accurate results in velocity and position, despite similar attitude estimation accuracy between both methods. Furthermore, the Chebyshev polynomial optimization demonstrates greater time efficiency, benefiting from the efficiency of Chebyshev polynomials and the additive nature of bias in the objective function, which allows for rapid bias correction even with large initial errors.

Table \ref{tab3} lists the statistical means of the RMSEs and averaged time costs for all segments in each sequence. The statistical means of the RMSEs are calculated as follows: 
\begin{equation}
    S M-R M S E=\sqrt{\frac{1}{H_s} \sum_{h=1}^{H_s} \sum_{k=1}^Z\left\|x_k^h-\hat{x}_k^h\right\|^2} \label{eq32}
\end{equation}
where $H_s$ represents the number of data segments. Consistent with the simulation results, the Chebyshev polynomial optimization outperforms preintegration in terms of velocity and position estimation accuracy. 
\begin{table}[htbp]
\centering
\caption{Statistical means of the RMSEs and averaged time costs for all segments in each sequence.}
\label{tab3}
\begin{tabular}{l|cccccc}
\toprule
 &  & \textbf{MH1} & \textbf{MH2} & \textbf{MH3} & \textbf{MH4} & \textbf{MH5} \\
\midrule
\multirow{4}{*}{\textbf{Chebyshev}} 
& Attitude(deg) & 0.2747 & \textbf{0.2602} & 0.2294 & 0.2881 & 0.1904 \\
& Velocity(m/s) & \textbf{0.1003} & \textbf{0.0872} & \textbf{0.1317} & \textbf{0.1063} & \textbf{0.0905} \\
& Position(m) & \textbf{0.0374} & \textbf{0.0317} & \textbf{0.0502} & \textbf{0.0376} & \textbf{0.0334} \\
& Averaged time cost(s) & \textbf{9.6616} & \textbf{9.6654} & \textbf{9.8676} & \textbf{9.1975} & \textbf{8.9111} \\
\midrule
\multirow{4}{*}{\textbf{Preintegration}} 
& Attitude(deg) & \textbf{0.2691} & 0.2784 & \textbf{0.2007} & \textbf{0.2571} & \textbf{0.1699} \\
& Velocity(m/s) & 0.1461 & 0.1070 & 0.2255 & 0.1526 & 0.4022 \\
& Position(m) & 0.0664 & 0.0635 & 0.0932 & 0.0863 & 0.1378 \\
& Averaged time cost(s) & 19.2917 & 16.9983 & 20.5341 & 19.8631 & 19.6818 \\
\bottomrule
\end{tabular}
\end{table}
Table \ref{tab4} details the specific improvement ratios corresponding to Table \ref{tab3}, showing varied degrees of improvement across different sequences – with velocity accuracy increasing by approximately 30\% and position accuracy by about 50\%. Moreover, the Chebyshev polynomial optimization achieves an approximate 50\% improvement in runtime efficiency. Because the state references for these sequences are derived from laser ranging and maximum likelihood estimation, the attitude truths are relatively less accurate. Thus, although some sequences show marginally inferior attitude estimation results of Chebyshev polynomial optimization as compared with preintegration, it is premature to conclude that the Chebyshev polynomial optimization's attitude estimation is inferior to preintegration.
\begin{table}[htbp]
\centering
\caption{The improvement ratios of the Chebyshev polynomial optimization relative to preintegration}
\label{tab4}
\begin{tabular}{lcccc}
\toprule
 & \textbf{Attitude (\%)} & \textbf{Velocity (\%)} & \textbf{Position (\%)} & \textbf{Time cost (\%)} \\
\midrule
\textbf{MH1} & -2.1 & \textbf{31.3} & \textbf{43.7} & \textbf{49.9} \\
\textbf{MH2} & \textbf{6.5} & \textbf{18.5} & \textbf{50.1} & \textbf{43.1} \\
\textbf{MH3} & -14.3 & \textbf{41.6} & \textbf{46.1} & \textbf{51.9} \\
\textbf{MH4} & -12.1 & \textbf{30.3} & \textbf{56.4} & \textbf{53.7} \\
\textbf{MH5} & -12.1 & \textbf{77.5} & \textbf{75.7} & \textbf{54.7} \\
\bottomrule
\end{tabular}
\end{table}

\section{Conclusion} \label{sec4}
The fusion of visual and inertial data for pose estimation is a key element in numerous applications. This paper proposes a continuous-time visual-inertial fusion algorithm based on Chebyshev polynomial optimization. By mapping poses as finite-order Chebyshev polynomials, the pose estimation problem is transformed into an optimization problem of polynomial coefficients. Extensive simulations and public dataset testing demonstrate that this method exhibits superior accuracy in batch optimization compared to preintegration. Specifically, there is an observed average decrease in velocity error by about 30\% and in position error by about 50\%. Additionally, this method shows a notable improvement in computational efficiency, with time cost reduced by nearly 50\% on average.

However, it is important to highlight the current limitations of this method. Firstly, the polynomial order selection in the algorithm is not yet adaptive, which could be a direction for further refinement. Secondly, the focus has been primarily on batch optimization; a corresponding sliding window approach has not yet been carried out, marking the current results as a crucial but initial step in this research area. Our future work will concentrate on overcoming these limitations by developing adaptive polynomial order selection and sliding window methods for Chebyshev polynomial optimization to enable real-time implementation and enhance practical applicability.

\section*{Declarations}

\begin{itemize}

\item Availability of data and materials

The datasets generated during and/or analysed during the current study are available in the EuRoC MAV repository, https://projects.asl.ethz.ch/datasets/doku.php?id=kmavvisualinertialdatasets.$^{[4]}$

\item Competing interests

The authors declare that they have no competing interests.

\item Funding

This paper was supported in part by National Key R\&D Program (2022YFB3903802), National Natural Science Foundation (62273228) and Postdoctoral Science Foundation and Fellowship Program of CPSF(GZC20231591,2023M732227).

\item Authors' contributions

HZ developed the method, wrote the paper and carried out the experiment; MZ and YW conceived the idea and reviewed the paper; QC assisted in the experimental evaluation and reviewed the paper. All authors read and approved the final manuscript.

\item Acknowledgements

Not applicable
\end{itemize}


\bibliography{sn-bibliography}

\begin{thebibliography}{}
\renewcommand{\doi}[1]{\url{https://doi.org/#1}}
\bibcommenthead

\bibitem [\protect \citeauthoryear {%
Berrut%
\ \BBA {} Klein%
}{%
Berrut%
\ \BBA {} Klein%
}{%
{\protect \APACyear {2014}}%
}]{%
bib31}
\APACinsertmetastar {%
bib31}%
\begin{APACrefauthors}%
Berrut, J\BHBI P.%
\BCBT {}\ \BBA {} Klein, G.%
\end{APACrefauthors}%
\unskip\
\newblock
\APACrefYearMonthDay{2014}{}{}.
\newblock
{\BBOQ}\APACrefatitle {Recent advances in linear barycentric rational interpolation} {Recent advances in linear barycentric rational interpolation}.{\BBCQ}
\newblock
\APACjournalVolNumPages{Journal of Computational and Applied Mathematics}{259}{}{95--107,}
\newblock

\newblock

\PrintBackRefs{\CurrentBib}

\bibitem [\protect \citeauthoryear {%
Bloesch%
, Burri%
, Omari%
, Hutter%
\BCBL {}\ \BBA {} Siegwart%
}{%
Bloesch%
\ \protect \BOthers {.}}{%
{\protect \APACyear {2017}}%
}]{%
bib4}
\APACinsertmetastar {%
bib4}%
\begin{APACrefauthors}%
Bloesch, M.%
, Burri, M.%
, Omari, S.%
, Hutter, M.%
\BCBL {} Siegwart, R.%
\end{APACrefauthors}%
\unskip\
\newblock
\APACrefYearMonthDay{2017}{}{}.
\newblock
{\BBOQ}\APACrefatitle {Iterated extended Kalman filter based visual-inertial odometry using direct photometric feedback} {Iterated extended kalman filter based visual-inertial odometry using direct photometric feedback}.{\BBCQ}
\newblock
\APACjournalVolNumPages{The International Journal of Robotics Research}{36}{10}{1053--1072,}
\newblock

\newblock

\PrintBackRefs{\CurrentBib}

\bibitem [\protect \citeauthoryear {%
Brossard%
, Bonnabel%
\BCBL {}\ \BBA {} Barrau%
}{%
Brossard%
\ \protect \BOthers {.}}{%
{\protect \APACyear {2018}}%
}]{%
bib6}
\APACinsertmetastar {%
bib6}%
\begin{APACrefauthors}%
Brossard, M.%
, Bonnabel, S.%
\BCBL {} Barrau, A.%
\end{APACrefauthors}%
\unskip\
\newblock
\APACrefYearMonthDay{2018}{}{}.
\newblock
{\BBOQ}\APACrefatitle {Invariant Kalman filtering for visual inertial SLAM} {Invariant kalman filtering for visual inertial slam}.{\BBCQ}
\newblock
 \APACrefbtitle {{2018 21st International Conference on Information Fusion (FUSION)}} {{2018 21st International Conference on Information Fusion (FUSION)}}\ (\BPGS\ 2021--2028).
\PrintBackRefs{\CurrentBib}

\bibitem [\protect \citeauthoryear {%
Burri%
\ \protect \BOthers {.}}{%
Burri%
\ \protect \BOthers {.}}{%
{\protect \APACyear {2016}}%
}]{%
bib32}
\APACinsertmetastar {%
bib32}%
\begin{APACrefauthors}%
Burri, M.%
, Nikolic, J.%
, Gohl, P.%
, Schneider, T.%
, Rehder, J.%
, Omari, S.%
\BDBL {}Siegwart, R.%
\end{APACrefauthors}%
\unskip\
\newblock
\APACrefYearMonthDay{2016}{}{}.
\newblock
{\BBOQ}\APACrefatitle {The EuRoC micro aerial vehicle datasets} {The euroc micro aerial vehicle datasets}.{\BBCQ}
\newblock
\APACjournalVolNumPages{The International Journal of Robotics Research}{35}{10}{1157--1163,}
\newblock

\newblock

\PrintBackRefs{\CurrentBib}

\bibitem [\protect \citeauthoryear {%
Campos%
, Elvira%
, Rodr{\'\i}guez%
, Montiel%
\BCBL {}\ \BBA {} Tard{\'o}s%
}{%
Campos%
\ \protect \BOthers {.}}{%
{\protect \APACyear {2021}}%
}]{%
bib11}
\APACinsertmetastar {%
bib11}%
\begin{APACrefauthors}%
Campos, C.%
, Elvira, R.%
, Rodr{\'\i}guez, J.J.G.%
, Montiel, J.M.%
\BCBL {} Tard{\'o}s, J.D.%
\end{APACrefauthors}%
\unskip\
\newblock
\APACrefYearMonthDay{2021}{}{}.
\newblock
{\BBOQ}\APACrefatitle {Orb-slam3: An accurate open-source library for visual, visual--inertial, and multimap slam} {Orb-slam3: An accurate open-source library for visual, visual--inertial, and multimap slam}.{\BBCQ}
\newblock
\APACjournalVolNumPages{IEEE Transactions on Robotics}{37}{6}{1874--1890,}
\newblock

\newblock

\PrintBackRefs{\CurrentBib}

\bibitem [\protect \citeauthoryear {%
Ebcin%
\ \BBA {} Veth%
}{%
Ebcin%
\ \BBA {} Veth%
}{%
{\protect \APACyear {2007}}%
}]{%
bib7}
\APACinsertmetastar {%
bib7}%
\begin{APACrefauthors}%
Ebcin, S.%
\BCBT {}\ \BBA {} Veth, M.%
\end{APACrefauthors}%
\unskip\
\newblock
\APACrefYearMonthDay{2007}{}{}.
\newblock
{\BBOQ}\APACrefatitle {Tightly-coupled image-aided inertial navigation using the unscented Kalman filter} {Tightly-coupled image-aided inertial navigation using the unscented kalman filter}.{\BBCQ}
\newblock
 \APACrefbtitle {{Proceedings of the 20th International Technical Meeting of the Satellite Division of The Institute of Navigation (ION GNSS 2007)}} {{Proceedings of the 20th International Technical Meeting of the Satellite Division of The Institute of Navigation (ION GNSS 2007)}}\ (\BPGS\ 1851--1860).
\PrintBackRefs{\CurrentBib}

\bibitem [\protect \citeauthoryear {%
Forster%
, Carlone%
, Dellaert%
\BCBL {}\ \BBA {} Scaramuzza%
}{%
Forster%
\ \protect \BOthers {.}}{%
{\protect \APACyear {2016}}%
}]{%
bib12}
\APACinsertmetastar {%
bib12}%
\begin{APACrefauthors}%
Forster, C.%
, Carlone, L.%
, Dellaert, F.%
\BCBL {} Scaramuzza, D.%
\end{APACrefauthors}%
\unskip\
\newblock
\APACrefYearMonthDay{2016}{}{}.
\newblock
{\BBOQ}\APACrefatitle {On-manifold preintegration for real-time visual--inertial odometry} {On-manifold preintegration for real-time visual--inertial odometry}.{\BBCQ}
\newblock
\APACjournalVolNumPages{IEEE Transactions on Robotics}{33}{1}{1--21,}
\newblock

\newblock

\PrintBackRefs{\CurrentBib}

\bibitem [\protect \citeauthoryear {%
Furgale%
, Barfoot%
\BCBL {}\ \BBA {} Sibley%
}{%
Furgale%
\ \protect \BOthers {.}}{%
{\protect \APACyear {2012}}%
}]{%
bib14}
\APACinsertmetastar {%
bib14}%
\begin{APACrefauthors}%
Furgale, P.%
, Barfoot, T.D.%
\BCBL {} Sibley, G.%
\end{APACrefauthors}%
\unskip\
\newblock
\APACrefYearMonthDay{2012}{}{}.
\newblock
{\BBOQ}\APACrefatitle {Continuous-time batch estimation using temporal basis functions} {Continuous-time batch estimation using temporal basis functions}.{\BBCQ}
\newblock
 \APACrefbtitle {{2012 IEEE International Conference on Robotics and Automation}} {{2012 IEEE International Conference on Robotics and Automation}}\ (\BPGS\ 2088--2095).
\PrintBackRefs{\CurrentBib}

\bibitem [\protect \citeauthoryear {%
Groves%
}{%
Groves%
}{%
{\protect \APACyear {2015}}%
}]{%
bib21}
\APACinsertmetastar {%
bib21}%
\begin{APACrefauthors}%
Groves, P.D.%
\end{APACrefauthors}%
\unskip\
\newblock
\APACrefYearMonthDay{2015}{}{}.
\newblock
{\BBOQ}\APACrefatitle {Principles of GNSS, inertial, and multisensor integrated navigation systems, [Book review]} {Principles of gnss, inertial, and multisensor integrated navigation systems, [book review]}.{\BBCQ}
\newblock
\APACjournalVolNumPages{IEEE Aerospace and Electronic Systems Magazine}{30}{2}{26--27,}
\newblock

\newblock

\PrintBackRefs{\CurrentBib}

\bibitem [\protect \citeauthoryear {%
Hartley%
\ \BBA {} Zisserman%
}{%
Hartley%
\ \BBA {} Zisserman%
}{%
{\protect \APACyear {2003}}%
}]{%
bib22}
\APACinsertmetastar {%
bib22}%
\begin{APACrefauthors}%
Hartley, R.%
\BCBT {}\ \BBA {} Zisserman, A.%
\end{APACrefauthors}%
\unskip\
\newblock
\APACrefYear{2003}.
\newblock
\APACrefbtitle {Multiple view geometry in computer vision} {Multiple view geometry in computer vision}.
\newblock
\APACaddressPublisher{}{{Cambridge university press}}.
\PrintBackRefs{\CurrentBib}

\bibitem [\protect \citeauthoryear {%
Huang%
}{%
Huang%
}{%
{\protect \APACyear {2019}}%
}]{%
bib1}
\APACinsertmetastar {%
bib1}%
\begin{APACrefauthors}%
Huang, G.%
\end{APACrefauthors}%
\unskip\
\newblock
\APACrefYearMonthDay{2019}{}{}.
\newblock
{\BBOQ}\APACrefatitle {Visual-Inertial Navigation: A Concise Review} {Visual-inertial navigation: A concise review}.{\BBCQ}
\newblock
 \APACrefbtitle {{2019 International Conference on Robotics and Automation (ICRA)}} {{2019 International Conference on Robotics and Automation (ICRA)}}\ (\BPG~9572-9582).
\PrintBackRefs{\CurrentBib}

\bibitem [\protect \citeauthoryear {%
Hug%
, B{\"a}nninger%
, Alzugaray%
\BCBL {}\ \BBA {} Chli%
}{%
Hug%
\ \protect \BOthers {.}}{%
{\protect \APACyear {2022}}%
}]{%
bib17}
\APACinsertmetastar {%
bib17}%
\begin{APACrefauthors}%
Hug, D.%
, B{\"a}nninger, P.%
, Alzugaray, I.%
\BCBL {} Chli, M.%
\end{APACrefauthors}%
\unskip\
\newblock
\APACrefYearMonthDay{2022}{}{}.
\newblock
{\BBOQ}\APACrefatitle {Continuous-time stereo-inertial odometry} {Continuous-time stereo-inertial odometry}.{\BBCQ}
\newblock
\APACjournalVolNumPages{IEEE Robotics and Automation Letters}{7}{3}{6455--6462,}
\newblock

\newblock

\PrintBackRefs{\CurrentBib}

\bibitem [\protect \citeauthoryear {%
Jazwinski%
}{%
Jazwinski%
}{%
{\protect \APACyear {2007}}%
}]{%
bib24}
\APACinsertmetastar {%
bib24}%
\begin{APACrefauthors}%
Jazwinski, A.H.%
\end{APACrefauthors}%
\unskip\
\newblock
\APACrefYear{2007}.
\newblock
\APACrefbtitle {Stochastic processes and filtering theory} {Stochastic processes and filtering theory}.
\newblock
\APACaddressPublisher{}{{Courier Corporation}}.
\PrintBackRefs{\CurrentBib}

\bibitem [\protect \citeauthoryear {%
Klein%
}{%
Klein%
}{%
{\protect \APACyear {2013}}%
}]{%
bib28}
\APACinsertmetastar {%
bib28}%
\begin{APACrefauthors}%
Klein, G.%
\end{APACrefauthors}%
\unskip\
\newblock
\APACrefYearMonthDay{2013}{}{}.
\newblock
{\BBOQ}\APACrefatitle {An extension of the Floater--Hormann family of barycentric rational interpolants} {An extension of the floater--hormann family of barycentric rational interpolants}.{\BBCQ}
\newblock
\APACjournalVolNumPages{Mathematics of Computation}{82}{284}{2273--2292,}
\newblock

\newblock

\PrintBackRefs{\CurrentBib}

\bibitem [\protect \citeauthoryear {%
Lang%
\ \protect \BOthers {.}}{%
Lang%
\ \protect \BOthers {.}}{%
{\protect \APACyear {2022}}%
}]{%
bib18}
\APACinsertmetastar {%
bib18}%
\begin{APACrefauthors}%
Lang, X.%
, Lv, J.%
, Huang, J.%
, Ma, Y.%
, Liu, Y.%
\BCBL {} Zuo, X.%
\end{APACrefauthors}%
\unskip\
\newblock
\APACrefYearMonthDay{2022}{}{}.
\newblock
{\BBOQ}\APACrefatitle {Ctrl-vio: Continuous-time visual-inertial odometry for rolling shutter cameras} {Ctrl-vio: Continuous-time visual-inertial odometry for rolling shutter cameras}.{\BBCQ}
\newblock
\APACjournalVolNumPages{IEEE Robotics and Automation Letters}{7}{4}{11537--11544,}
\newblock

\newblock

\PrintBackRefs{\CurrentBib}

\bibitem [\protect \citeauthoryear {%
Leutenegger%
, Lynen%
, Bosse%
, Siegwart%
\BCBL {}\ \BBA {} Furgale%
}{%
Leutenegger%
\ \protect \BOthers {.}}{%
{\protect \APACyear {2015}}%
}]{%
bib8}
\APACinsertmetastar {%
bib8}%
\begin{APACrefauthors}%
Leutenegger, S.%
, Lynen, S.%
, Bosse, M.%
, Siegwart, R.%
\BCBL {} Furgale, P.%
\end{APACrefauthors}%
\unskip\
\newblock
\APACrefYearMonthDay{2015}{}{}.
\newblock
{\BBOQ}\APACrefatitle {Keyframe-based visual--inertial odometry using nonlinear optimization} {Keyframe-based visual--inertial odometry using nonlinear optimization}.{\BBCQ}
\newblock
\APACjournalVolNumPages{The International Journal of Robotics Research}{34}{3}{314--334,}
\newblock

\newblock

\PrintBackRefs{\CurrentBib}

\bibitem [\protect \citeauthoryear {%
Li%
\ \BBA {} Mourikis%
}{%
Li%
\ \BBA {} Mourikis%
}{%
{\protect \APACyear {2012}}%
}]{%
bib3}
\APACinsertmetastar {%
bib3}%
\begin{APACrefauthors}%
Li, M.%
\BCBT {}\ \BBA {} Mourikis, A.I.%
\end{APACrefauthors}%
\unskip\
\newblock
\APACrefYearMonthDay{2012}{}{}.
\newblock
{\BBOQ}\APACrefatitle {Improving the accuracy of EKF-based visual-inertial odometry} {Improving the accuracy of ekf-based visual-inertial odometry}.{\BBCQ}
\newblock
 \APACrefbtitle {{2012 IEEE International Conference on Robotics and Automation}} {{2012 IEEE International Conference on Robotics and Automation}}\ (\BPGS\ 828--835).
\PrintBackRefs{\CurrentBib}

\bibitem [\protect \citeauthoryear {%
Lovegrove%
, Patron-Perez%
\BCBL {}\ \BBA {} Sibley%
}{%
Lovegrove%
\ \protect \BOthers {.}}{%
{\protect \APACyear {2013}}%
}]{%
bib16}
\APACinsertmetastar {%
bib16}%
\begin{APACrefauthors}%
Lovegrove, S.%
, Patron-Perez, A.%
\BCBL {} Sibley, G.%
\end{APACrefauthors}%
\unskip\
\newblock
\APACrefYearMonthDay{2013}{}{}.
\newblock
{\BBOQ}\APACrefatitle {Spline Fusion: A continuous-time representation for visual-inertial fusion with application to rolling shutter cameras.} {Spline fusion: A continuous-time representation for visual-inertial fusion with application to rolling shutter cameras.}{\BBCQ}
\newblock
 \APACrefbtitle {{BMVC}} {{BMVC}}\ (\BVOL~2, \BPG~8).
\PrintBackRefs{\CurrentBib}

\bibitem [\protect \citeauthoryear {%
Ma%
, Soatto%
, Ko{\v{s}}eck{\'a}%
\BCBL {}\ \BBA {} Sastry%
}{%
Ma%
\ \protect \BOthers {.}}{%
{\protect \APACyear {2004}}%
}]{%
bib23}
\APACinsertmetastar {%
bib23}%
\begin{APACrefauthors}%
Ma, Y.%
, Soatto, S.%
, Ko{\v{s}}eck{\'a}, J.%
\BCBL {} Sastry, S.%
\end{APACrefauthors}%
\unskip\
\newblock
\APACrefYear{2004}.
\newblock
\APACrefbtitle {An invitation to 3-d vision: from images to geometric models} {An invitation to 3-d vision: from images to geometric models}\ (\BVOL~26).
\newblock
\APACaddressPublisher{}{Springer}.
\PrintBackRefs{\CurrentBib}

\bibitem [\protect \citeauthoryear {%
Mo%
\ \BBA {} Sattar%
}{%
Mo%
\ \BBA {} Sattar%
}{%
{\protect \APACyear {2022}}%
}]{%
bib15}
\APACinsertmetastar {%
bib15}%
\begin{APACrefauthors}%
Mo, J.%
\BCBT {}\ \BBA {} Sattar, J.%
\end{APACrefauthors}%
\unskip\
\newblock
\APACrefYearMonthDay{2022}{}{}.
\newblock
{\BBOQ}\APACrefatitle {Continuous-time spline visual-inertial odometry} {Continuous-time spline visual-inertial odometry}.{\BBCQ}
\newblock
 \APACrefbtitle {{2022 International Conference on Robotics and Automation (ICRA)}} {{2022 International Conference on Robotics and Automation (ICRA)}}\ (\BPGS\ 9492--9498).
\PrintBackRefs{\CurrentBib}

\bibitem [\protect \citeauthoryear {%
Mourikis%
\ \BBA {} Roumeliotis%
}{%
Mourikis%
\ \BBA {} Roumeliotis%
}{%
{\protect \APACyear {2007}}%
}]{%
bib2}
\APACinsertmetastar {%
bib2}%
\begin{APACrefauthors}%
Mourikis, A.I.%
\BCBT {}\ \BBA {} Roumeliotis, S.I.%
\end{APACrefauthors}%
\unskip\
\newblock
\APACrefYearMonthDay{2007}{}{}.
\newblock
{\BBOQ}\APACrefatitle {A Multi-State Constraint Kalman Filter for Vision-aided Inertial Navigation} {A multi-state constraint kalman filter for vision-aided inertial navigation}.{\BBCQ}
\newblock
 \APACrefbtitle {{Proceedings 2007 IEEE International Conference on Robotics and Automation}} {{Proceedings 2007 IEEE International Conference on Robotics and Automation}}\ (\BPG~3565-3572).
\PrintBackRefs{\CurrentBib}

\bibitem [\protect \citeauthoryear {%
Press%
}{%
Press%
}{%
{\protect \APACyear {2007}}%
}]{%
bib27}
\APACinsertmetastar {%
bib27}%
\begin{APACrefauthors}%
Press, W.H.%
\end{APACrefauthors}%
\unskip\
\newblock
\APACrefYear{2007}.
\newblock
\APACrefbtitle {Numerical recipes 3rd edition: The art of scientific computing} {Numerical recipes 3rd edition: The art of scientific computing}.
\newblock
\APACaddressPublisher{}{Cambridge university press}.
\PrintBackRefs{\CurrentBib}

\bibitem [\protect \citeauthoryear {%
K.~Qin%
}{%
K.~Qin%
}{%
{\protect \APACyear {1998}}%
}]{%
bib19}
\APACinsertmetastar {%
bib19}%
\begin{APACrefauthors}%
Qin, K.%
\end{APACrefauthors}%
\unskip\
\newblock
\APACrefYearMonthDay{1998}{}{}.
\newblock
{\BBOQ}\APACrefatitle {General matrix representations for B-splines} {General matrix representations for b-splines}.{\BBCQ}
\newblock
 \APACrefbtitle {{Proceedings Pacific Graphics' 98. Sixth Pacific Conference on Computer Graphics and Applications (Cat. No. 98EX208)}} {{Proceedings Pacific Graphics' 98. Sixth Pacific Conference on Computer Graphics and Applications (Cat. No. 98EX208)}}\ (\BPGS\ 37--43).
\PrintBackRefs{\CurrentBib}

\bibitem [\protect \citeauthoryear {%
T.~Qin%
, Li%
\BCBL {}\ \BBA {} Shen%
}{%
T.~Qin%
\ \protect \BOthers {.}}{%
{\protect \APACyear {2018}}%
}]{%
bib9}
\APACinsertmetastar {%
bib9}%
\begin{APACrefauthors}%
Qin, T.%
, Li, P.%
\BCBL {} Shen, S.%
\end{APACrefauthors}%
\unskip\
\newblock
\APACrefYearMonthDay{2018}{}{}.
\newblock
{\BBOQ}\APACrefatitle {Vins-mono: A robust and versatile monocular visual-inertial state estimator} {Vins-mono: A robust and versatile monocular visual-inertial state estimator}.{\BBCQ}
\newblock
\APACjournalVolNumPages{IEEE Transactions on Robotics}{34}{4}{1004--1020,}
\newblock

\newblock

\PrintBackRefs{\CurrentBib}

\bibitem [\protect \citeauthoryear {%
T.~Qin%
\ \BBA {} Shen%
}{%
T.~Qin%
\ \BBA {} Shen%
}{%
{\protect \APACyear {{\protect \bibnodate {}}}}%
}]{%
bib13}
\APACinsertmetastar {%
bib13}%
\begin{APACrefauthors}%
Qin, T.%
\BCBT {}\ \BBA {} Shen, S.%
\end{APACrefauthors}%
\unskip\
\newblock
\APACrefYearMonthDay{{\protect \bibnodate {}}}{}{}.
\newblock
{\BBOQ}\APACrefatitle {Online temporal calibration for monocular visual-inertial systems. In 2018 {IEEE}} {Online temporal calibration for monocular visual-inertial systems. in 2018 {IEEE}}.{\BBCQ}
\newblock
 \APACrefbtitle {{RSJ International Conference on Intelligent Robots and Systems (IROS)}} {{RSJ International Conference on Intelligent Robots and Systems (IROS)}}\ (\BPGS\ 3662--3669).
\PrintBackRefs{\CurrentBib}

\bibitem [\protect \citeauthoryear {%
Ross%
\ \BBA {} Karpenko%
}{%
Ross%
\ \BBA {} Karpenko%
}{%
{\protect \APACyear {2012}}%
}]{%
bib29}
\APACinsertmetastar {%
bib29}%
\begin{APACrefauthors}%
Ross, I.M.%
\BCBT {}\ \BBA {} Karpenko, M.%
\end{APACrefauthors}%
\unskip\
\newblock
\APACrefYearMonthDay{2012}{}{}.
\newblock
{\BBOQ}\APACrefatitle {A review of pseudospectral optimal control: From theory to flight} {A review of pseudospectral optimal control: From theory to flight}.{\BBCQ}
\newblock
\APACjournalVolNumPages{Annual Reviews in Control}{36}{2}{182--197,}
\newblock

\newblock

\PrintBackRefs{\CurrentBib}

\bibitem [\protect \citeauthoryear {%
Strang%
}{%
Strang%
}{%
{\protect \APACyear {2020}}%
}]{%
bib30}
\APACinsertmetastar {%
bib30}%
\begin{APACrefauthors}%
Strang, G.%
\end{APACrefauthors}%
\unskip\
\newblock
\APACrefYearMonthDay{2020}{}{}.
\newblock
{\BBOQ}\APACrefatitle {Introduction to Applied Linear Algebra: Vectors, Matrices, and Least Squares [Bookshelf]} {Introduction to applied linear algebra: Vectors, matrices, and least squares [bookshelf]}.{\BBCQ}
\newblock
\APACjournalVolNumPages{IEEE Control Systems Magazine}{40}{6}{136--136,}
\newblock

\newblock

\PrintBackRefs{\CurrentBib}

\bibitem [\protect \citeauthoryear {%
Thrun%
}{%
Thrun%
}{%
{\protect \APACyear {2002}}%
}]{%
bib5}
\APACinsertmetastar {%
bib5}%
\begin{APACrefauthors}%
Thrun, S.%
\end{APACrefauthors}%
\unskip\
\newblock
\APACrefYearMonthDay{2002}{}{}.
\newblock
{\BBOQ}\APACrefatitle {Probabilistic robotics} {Probabilistic robotics}.{\BBCQ}
\newblock
\APACjournalVolNumPages{Communications of the ACM}{45}{3}{52--57,}
\newblock

\newblock

\PrintBackRefs{\CurrentBib}

\bibitem [\protect \citeauthoryear {%
Trefethen%
}{%
Trefethen%
}{%
{\protect \APACyear {2019}}%
}]{%
bib25}
\APACinsertmetastar {%
bib25}%
\begin{APACrefauthors}%
Trefethen, L.N.%
\end{APACrefauthors}%
\unskip\
\newblock
\APACrefYear{2019}.
\newblock
\APACrefbtitle {Approximation Theory and Approximation Practice, Extended Edition} {Approximation theory and approximation practice, extended edition}.
\newblock
\APACaddressPublisher{}{{SIAM}}.
\PrintBackRefs{\CurrentBib}

\bibitem [\protect \citeauthoryear {%
Von~Stumberg%
, Usenko%
\BCBL {}\ \BBA {} Cremers%
}{%
Von~Stumberg%
\ \protect \BOthers {.}}{%
{\protect \APACyear {2018}}%
}]{%
bib10}
\APACinsertmetastar {%
bib10}%
\begin{APACrefauthors}%
Von~Stumberg, L.%
, Usenko, V.%
\BCBL {} Cremers, D.%
\end{APACrefauthors}%
\unskip\
\newblock
\APACrefYearMonthDay{2018}{}{}.
\newblock
{\BBOQ}\APACrefatitle {Direct sparse visual-inertial odometry using dynamic marginalization} {Direct sparse visual-inertial odometry using dynamic marginalization}.{\BBCQ}
\newblock
 \APACrefbtitle {{2018 IEEE International Conference on Robotics and Automation (ICRA)}} {{2018 IEEE International Conference on Robotics and Automation (ICRA)}}\ (\BPGS\ 2510--2517).
\PrintBackRefs{\CurrentBib}

\bibitem [\protect \citeauthoryear {%
Wu%
}{%
Wu%
}{%
{\protect \APACyear {2019}}%
}]{%
bib26}
\APACinsertmetastar {%
bib26}%
\begin{APACrefauthors}%
Wu, Y.%
\end{APACrefauthors}%
\unskip\
\newblock
\APACrefYearMonthDay{2019}{}{}.
\newblock
{\BBOQ}\APACrefatitle {iNavFIter: Next-generation inertial navigation computation based on functional iteration} {inavfiter: Next-generation inertial navigation computation based on functional iteration}.{\BBCQ}
\newblock
\APACjournalVolNumPages{IEEE Transactions on Aerospace and Electronic Systems}{56}{3}{2061--2082,}
\newblock

\newblock

\PrintBackRefs{\CurrentBib}

\bibitem [\protect \citeauthoryear {%
Zhu%
\ \BBA {} Wu%
}{%
Zhu%
\ \BBA {} Wu%
}{%
{\protect \APACyear {2022}}%
}]{%
bib20}
\APACinsertmetastar {%
bib20}%
\begin{APACrefauthors}%
Zhu, M.%
\BCBT {}\ \BBA {} Wu, Y.%
\end{APACrefauthors}%
\unskip\
\newblock
\APACrefYearMonthDay{2022}{}{}.
\newblock
{\BBOQ}\APACrefatitle {ChevOpt: Continuous-Time State Estimation by Chebyshev Polynomial Optimization} {Chevopt: Continuous-time state estimation by chebyshev polynomial optimization}.{\BBCQ}
\newblock
\APACjournalVolNumPages{IEEE Transactions on Signal Processing}{70}{}{3136--3147,}
\newblock

\newblock

\PrintBackRefs{\CurrentBib}

\bibitem [\protect \citeauthoryear {%
Zhu%
\ \BBA {} Wu%
}{%
Zhu%
\ \BBA {} Wu%
}{%
{\protect \APACyear {2023}}%
}]{%
bib33}
\APACinsertmetastar {%
bib33}%
\begin{APACrefauthors}%
Zhu, M.%
\BCBT {}\ \BBA {} Wu, Y.%
\end{APACrefauthors}%
\unskip\
\newblock
\APACrefYearMonthDay{2023}{}{}.
\newblock
{\BBOQ}\APACrefatitle {Inertial-Based Navigation by Polynomial Optimization: Inertial-Magnetic Attitude Estimation} {Inertial-based navigation by polynomial optimization: Inertial-magnetic attitude estimation}.{\BBCQ}
\newblock
\APACjournalVolNumPages{IEEE Transactions on Aerospace and Electronic Systems}{59}{6}{7772-7783,}
\newblock
\begin{APACrefDOI} \doi{10.1109/TAES.2023.3294176} \end{APACrefDOI}
\newblock

\newblock

\PrintBackRefs{\CurrentBib}

\end{thebibliography}
\end{document}